\documentclass[sigconf]{acmart}
\usepackage{amsthm}
\usepackage{subcaption}
\usepackage{paralist}
\usepackage{bm}
\usepackage{amsfonts}
\usepackage{cases}
\usepackage{booktabs}
\usepackage{threeparttable}
\usepackage{epstopdf}
\usepackage{upgreek}
\usepackage{endnotes}
\usepackage{etoolbox}
\usepackage{algpseudocode}
\usepackage{xspace}
\usepackage{array}
\usepackage{enumitem}
\usepackage{hyperref}
\usepackage{balance}
\usepackage{multirow}
\usepackage{amsmath}
\usepackage{xcolor}
\usepackage{url}
\usepackage{graphicx}
\usepackage[ruled,linesnumbered]{algorithm2e}

\newcommand{\eat}[1]{}

\AtBeginDocument{%
  \providecommand\BibTeX{{%
    \normalfont B\kern-0.5em{\scshape i\kern-0.25em b}\kern-0.8em\TeX}}}

\newcommand{\ie}{\emph{i.e.,}\xspace}
\newcommand{\eg}{\emph{e.g.,}\xspace}
\newcommand{\etc}{\emph{etc.}\xspace}
\newcommand{\etal}{\emph{et al.}\xspace}

\newcommand{\baby}{\textsc{Meta-Pec}\xspace}

\copyrightyear{2023}
\acmYear{2023}
\setcopyright{acmlicensed}
\acmConference[KDD '23] {Proceedings of the 29th ACM SIGKDD Conference on Knowledge Discovery and Data Mining}{August 6--10, 2023}{Long Beach, CA, USA.}
\acmBooktitle{Proceedings of the 29th ACM SIGKDD Conference on Knowledge Discovery and Data Mining (KDD '23), August 6--10, 2023, Long Beach, CA, USA}
\acmPrice{15.00}
\acmISBN{979-8-4007-0103-0/23/08}
\acmDOI{10.1145/3580305.3599767}

\begin{document}
\title{A Preference-aware Meta-optimization Framework for Personalized Vehicle Energy Consumption Estimation}
\thanks{$^*$ Corresponding author.}
\author{Siqi Lai}
\affiliation{\institution{The Hong Kong University of Science and Technology (Guangzhou)}}
\email{siqilai@hkust-gz.edu.cn}

\author{Weijia Zhang}
\affiliation{\institution{The Hong Kong University of Science and Technology (Guangzhou)}}
\email{wzhang411@connect.hkust-gz.edu.cn}

\author{Hao Liu$^*$}
\affiliation{
\institution{The Hong Kong University of Science and Technology (Guangzhou)}
\institution{The Hong Kong University of Science and Technology}
}
\email{liuh@ust.hk}
\renewcommand{\shortauthors}{Siqi Lai, Weijia Zhang, \& Hao Liu}

\makeatletter
\def\@ACM@checkaffil{
    \if@ACM@instpresent\else
    \ClassWarningNoLine{\@classname}{No institution present for an affiliation}%
    \fi
    \if@ACM@citypresent\else
    \ClassWarningNoLine{\@classname}{No city present for an affiliation}%
    \fi
    \if@ACM@countrypresent\else
        \ClassWarningNoLine{\@classname}{No country present for an affiliation}%
    \fi
}
\makeatother

\begin{abstract}
Vehicle Energy Consumption (VEC) estimation aims to predict the total energy required for a given trip before it starts, which is of great importance to trip planning and transportation sustainability.
Existing approaches mainly focus on extracting statistically significant factors from typical trips to improve the VEC estimation.
However, the energy consumption of each vehicle may diverge widely due to the personalized driving behavior under varying travel contexts.
To this end, this paper proposes a preference-aware meta-optimization framework (\baby) for personalized vehicle energy consumption estimation. 
Specifically, we first propose a spatiotemporal behavior learning module to capture the latent driver preference hidden in historical trips.
Moreover, based on the memorization of driver preference, we devise a selection-based driving behavior prediction module to infer driver-specific driving patterns on a given route, which provides additional basis and supervision signals for VEC estimation.
Besides, a driver-specific meta-optimization scheme is proposed to enable fast model adaption by learning and sharing transferable knowledge globally.
Extensive experiments on two real-world datasets show the superiority of our proposed framework against ten numerical and data-driven machine learning baselines. 
The source code is available at \url{https://github.com/usail-hkust/Meta-Pec}.
\eat{
Vehicle energy consumption (EC) estimation aims to predict the trip's fuel (or energy) consumption before it starts. This task has drawn a lot of attention due to its success in helping drivers to plan their future trips and forecasting vehicle exhaustion. Previous EC estimation models pervasively focused on identifying the most significant factors of EC, then utilizing their statistical data to make the estimation, which failed to explore in-depth driver-specific information. This paper proposes a novel \textbf{Meta}-optimization-based Model for \textbf{P}ersonalized Vehicle \textbf{E}nergy \textbf{C}onsumption (\baby) estimation to tackle this problem.

Specifically, we developed a spatial-temporal-behavioral learning module to effectively extract driver driving preferences hidden in historical trips. Moreover, based on this personalization information, we design a select-based driver behavior decoder to predict how the driver will drive through the planned route, providing extra EC estimation basis and additional supervision signals. Finally, we also introduce a meta-optimization algorithm to learn globally shared knowledge and enable the model to fast adapt to every driver by fine-tuning. Experiments conducted on two real-world datasets show our proposed model surpasses all the baselines and achieves state-of-the-art.}
\vspace{-7pt}
\end{abstract}



\maketitle


\eat{
\begin{figure}[htbp]
  \begin{subfigure}{0.15\textwidth}
    \centering
    \includegraphics[width=0.95\linewidth]{figures/3_bold.png}
    \caption{Trip 1: Driver 1.}
    \label{fig:driver11}
  \end{subfigure}
  \begin{subfigure}{0.15\textwidth}
    \centering
    \includegraphics[width=0.95\linewidth]{figures/5_bold.png}
    \caption{Trip 2: Driver 1.}
    \label{fig:driver12}
  \end{subfigure}
  \begin{subfigure}{0.15\textwidth}
    \centering
    \includegraphics[width=0.95\linewidth]{figures/driver2.png}
    \caption{Trip 3: Driver 2.}
    \label{fig:driver2}
  \end{subfigure}
  \caption{Different energy usage patterns on the same roads.}
  \label{fig:sameroad}
\end{figure}
}

\section{Introduction}
\begin{figure}[t]
\setlength{\abovecaptionskip}{0.3cm}
\setlength{\belowcaptionskip}{-0.2cm}
\centering
\includegraphics[width=0.5\textwidth]{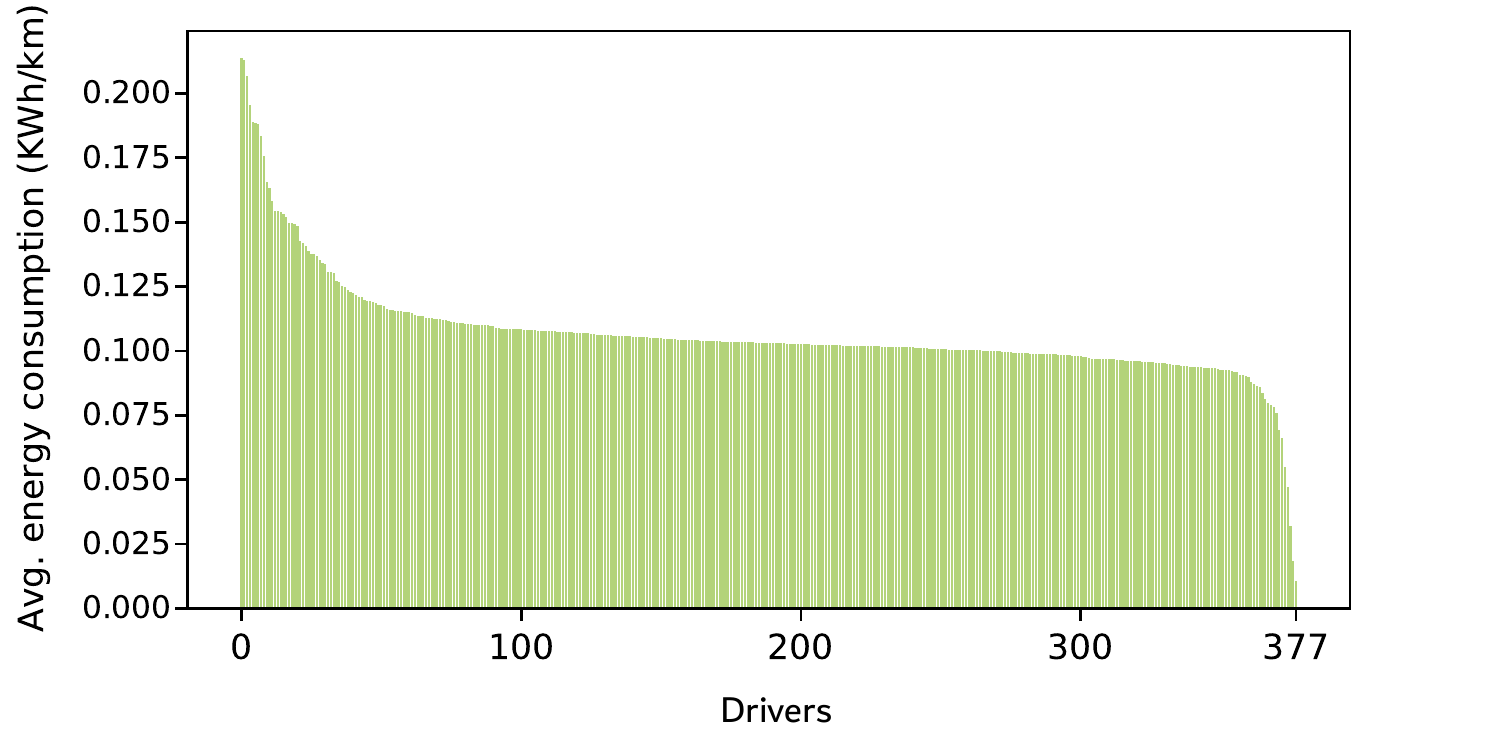}
\centering
\caption{The driver-varying energy consumption distribution of a real-world dataset in Shenzhen, all 377 vehicles are of the same type.}
\label{fig:intro}
\end{figure}

The rapid transportation network expansion and traffic demand growth raise public concerns about the efficiency, sustainability, and resilience of the urban transportation system.
From 2015 to 2030, as reported by the United Nations\footnote{https://www.un.org/sustainabledevelopment/progress-report/}, the number of vehicles on the road is approximately double, and global traffic is likely to increase by $50\%$.
Therefore, the accurate estimation of Vehicle Energy Consumption (VEC) is of great importance to the decision-making of urban governance and travel planning \cite{rauh2015understanding}. 

Prior studies on VEC estimation can be roughly divided into two categories: 
the \emph{Numerical methods}~\cite{de2015energy, ojeda2017fuel, ding2017greenplanner} and the \emph{Data-driven methods}~\cite{liu2021ldferr, elmi2021deepfec, petkevicius2021probabilistic, hua2022fine, moawad2021deep}. 
Specifically, the numerical methods mainly focus on identifying factors that have the most significant influence on VEC. 
For example, Y. Al-Wreikat \etal \cite{al2021driving} propose to partition drivers into multiple classes and quantify the average VECs. 
The data-driven methods, on the other hand, aim to automatically extract and utilize relevant knowledge by leveraging machine learning tools, \eg linear regression (LR)~\cite{de2017data}, decision tree (DT)~\cite{roy2022reliable}, support vector machine (SVM)~\cite{perrotta2017application}, \etc
Inspired by the recent advances of deep learning, deep neural networks such as Long-Short-Term Memory (LSTM) \cite{chen2021data} and Transformer~\cite{wang2022personalized} have also been adopted to analyze road-level energy consumption.
However, existing approaches make predictions based on handcrafted statistical features, which overlook the personalized nature of varying driving behaviors under different travel contexts.

In this work, we investigate the personalized vehicle energy consumption estimation problem with a consideration of fine-grained individual driving preferences hidden in historical trajectories. 
However, three major challenges arise towards this goal.
First, the driving preference describes the driver's long-term intention~(\eg overtaking, braking, changing lanes under different travel contexts), which is critical to the overall energy consumption.
As depicted in Figure~\ref{fig:intro}, the average energy consumption of different drivers may vary significantly, indicating the necessity of incorporating driver preference for VEC estimation.
However, the existing approach~\cite{wang2022personalized} represents latent driver preference via handcrafted features, which lead to lossy driving preference preservation.
Thus, how to capture the latent driving preference in an effective way is the first challenge.
Second, the behavior of a driver under different travel contexts may also vary. Estimating the intentional driving behavior of a given route can provide additional signals to guide the learning direction of the prediction model. However, the historical trajectory is noisy and may in large-scale.
How to quantify the personalized driving behavior on a target route based on historical data in a cost-effective manner is another challenge.
Third, although a unified model can provide predictions for all drivers, the estimation for drivers with only a few trajectories may be biased and error-prone.
The last challenge is how to share transferable knowledge between drivers so that to benefit the long-tail prediction.

To this end, in this paper, we propose a preference-aware meta-optimization framework, \baby, to deliver more effective personalized vehicle energy consumption estimation.
Specifically, we first propose a driving preference learning module to capture the latent spatiotemporal preference of each driver hidden in high-dimensional historical trajectories in an end-to-end manner.
Moreover, we devise a selection-based driving behavior prediction module to estimate the future behaviors of a driver on a given route.
In particular, the predicted behaviors provide additional supervision signals for model learning by incorporating the information from similar historical trips.
Furthermore, we propose a driver-specific meta-optimization scheme to allow fast model adaption to data-insufficient drivers, where the transferable knowledge is encoded in a global parameter initialization.

In summary, the major contributions of this paper are as follows.
(1) We investigate the personalized vehicle energy consumption estimation problem, which is beneficial to various downstream applications, such as fuel-efficient trip planning and sustainable urban transportation system policy-making.
(2) We propose a preference-aware meta-optimization framework to incorporate the latent driving behavior knowledge hidden in past trajectories. To our knowledge, this is the first work that utilizes meta-learning to tackle the cold-start problem in the VEC estimation task.
(3) Extensive experiments on two real-world datasets demonstrate the superiority of \baby compared with ten numerical and data-driven state-of-the-art approaches.

\section{PRELIMINARIES}\label{sec:prelim}

This section introduces some important definitions and the formal problem statement.

\textbf{Road network}. The road network consists of a set of road segments  $E = \{e_1, e_2, \dots, e_N\}$ and intersection joints, where $N$ is the number of road segments in the city. We use $\mathbf{x}^e_i$ to denote the road segment $i$'s features~(\eg length, number of lanes). 

\textbf{Route}. A route $R=[e_1, e_2, \dots, e_n]$ is a road segment sequence that a vehicle will traverse, where $n$ is the number of road segments. We denote $R_c$ as the route of the target trip and $R_i$ ($i \in \mathbb{N}^+$) as the route of a historical trip.

\textbf{Trajectory}. A trajectory $T_i=[(p_j, t_j, \mathbf{x}^l_j, y_j)]^m_{j=1}$ is a sequence of sample points logged by GPS devices, where $p_j$ is the distance the driver has traveled from the origin to the current location ($p_1=0$), $t_j$ is the time elapsed since the trip started ($t_1=0$), $\mathbf{x}^l_j$ are features describing the current state of the vehicle (\eg speed, acceleration), $y_j$ is the energy consumption from the origin to the current point $j$ ($y_1=0$), and $m$ is the number of sample points. We take $y=y_m$ as the ground truth label of the total energy consumption of the trip.

\textbf{Historical trips}. A driver $u$'s historical trips is defined as $H^u=\{(R_i, T_i)\}^M_{i=1} $, where $R_i$ is the route, $T_i$ is the corresponding trajectory, and $M$ is the total number of historical trips.

\textbf{Problem definition}. We define personalized vehicle energy consumption estimation as a supervised-learning task.
Given a target route $R_c$ and a driver $u$, we aim to predict the energy consumption of the trip based on $u$'s historical trips $H^u$,
\begin{align}
    f_\theta: (R_c, H^u) \rightarrow y,
\end{align}
where $f_\theta$ is the model parameterized by $\theta$ that we aim to learn,  $y$ is the ground truth energy consumption.

\section{Data Description and Analysis}\label{sec:data}

In this section, we describe the datasets used for \baby with a primary data analysis.

\begin{table}[tb]
 \small
 \centering
 \caption{Statistics of datasets.}
  \begin{tabular}{lll}
   \toprule
   \textbf{Category} & \textbf{VED} & \textbf{ETTD}\\
   \midrule
   \# of trips & 25,661 & 18,546 \\
   \# of drivers & 348 & 377 \\
   Time span & 11/1/2017-11/7/2018 & 10/22/2014 \\
   \midrule
   \# of total distance & 130,864 km & 87,251 km \\
   Avg. route length & 4.81 km & 5.10 km\\
   Avg. speed & 38.59 km/h & 13.54 km/h \\
   \bottomrule
 \end{tabular}
 \label{tab:dataStats}
\end{table}

\begin{figure*}[!]
  \begin{subfigure}{0.24\textwidth}
    \centering
    \includegraphics[width=\linewidth]{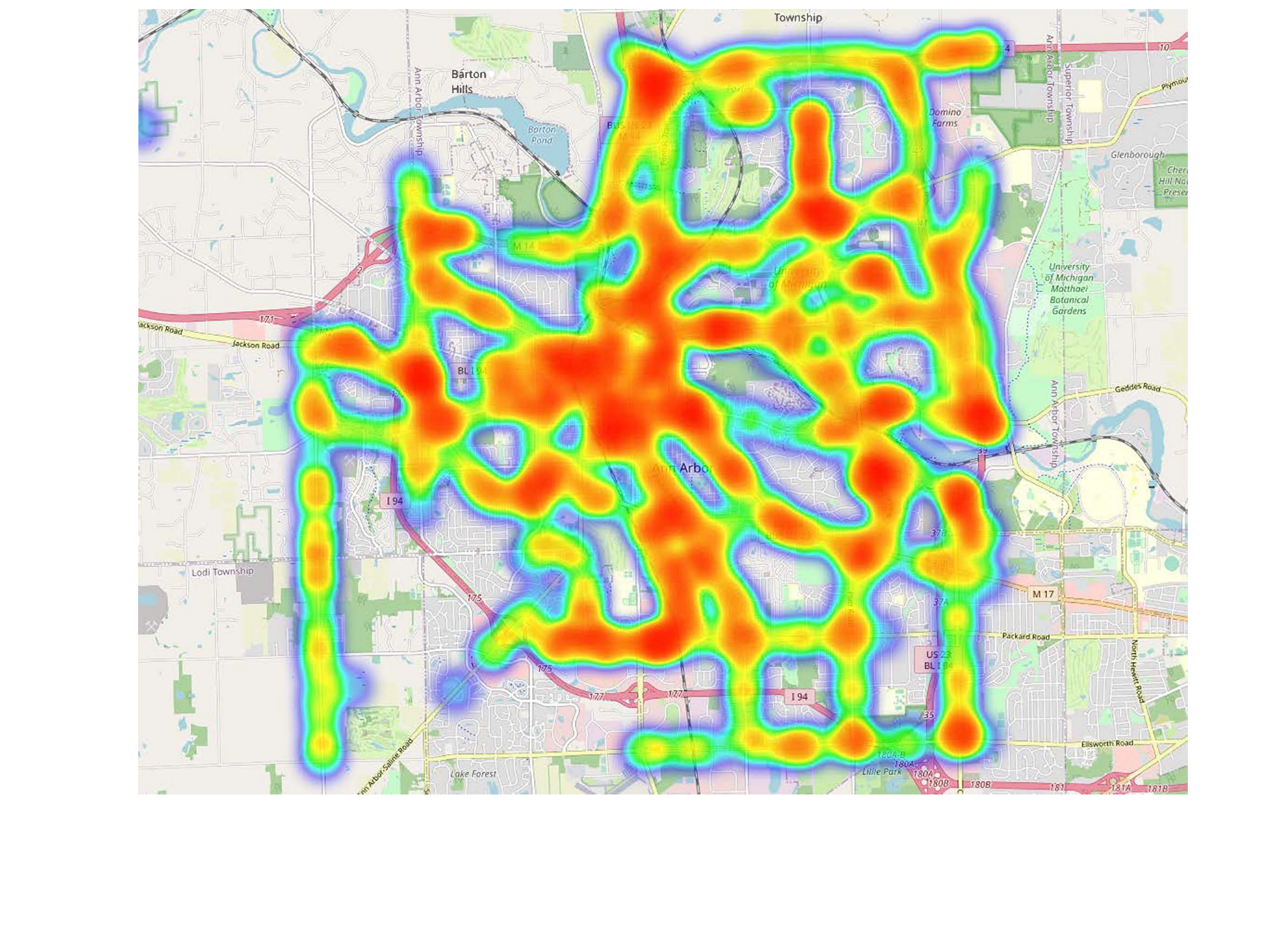}
    \caption{Spatial distribution.}
    \label{fig:evenergy}
  \end{subfigure}
  \begin{subfigure}{0.24\textwidth}
    \centering
    \includegraphics[width=\linewidth]{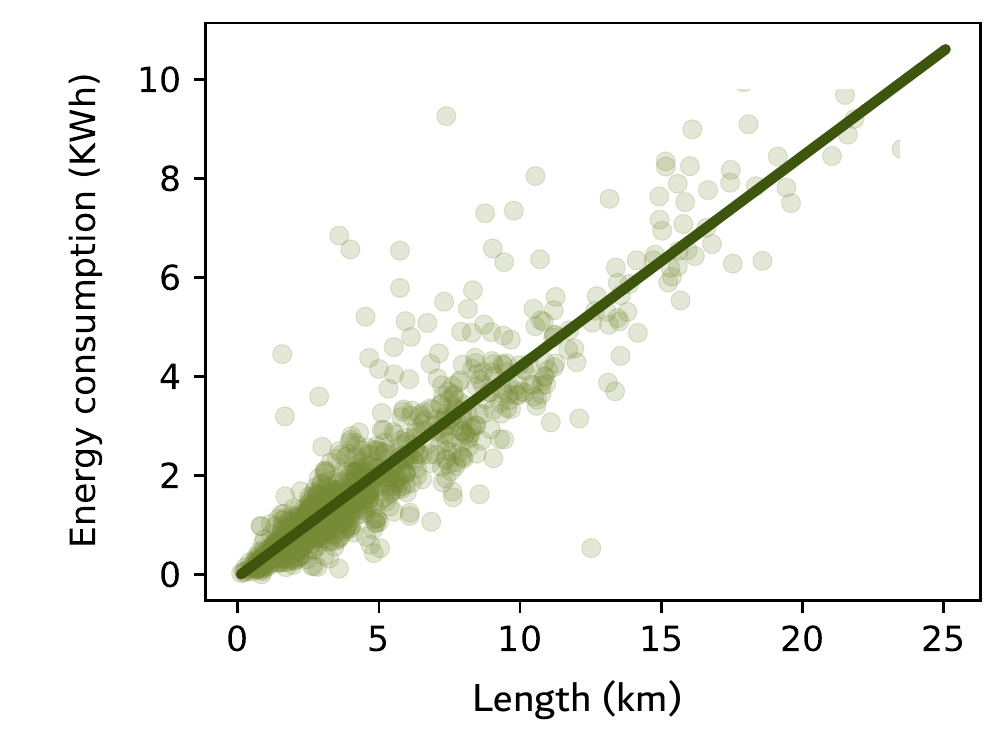}
    \caption{Trip length distribution.}
    \label{fig:distanceenergy}
  \end{subfigure}
  \begin{subfigure}{0.24\textwidth}
    \centering
    \includegraphics[width=\linewidth]{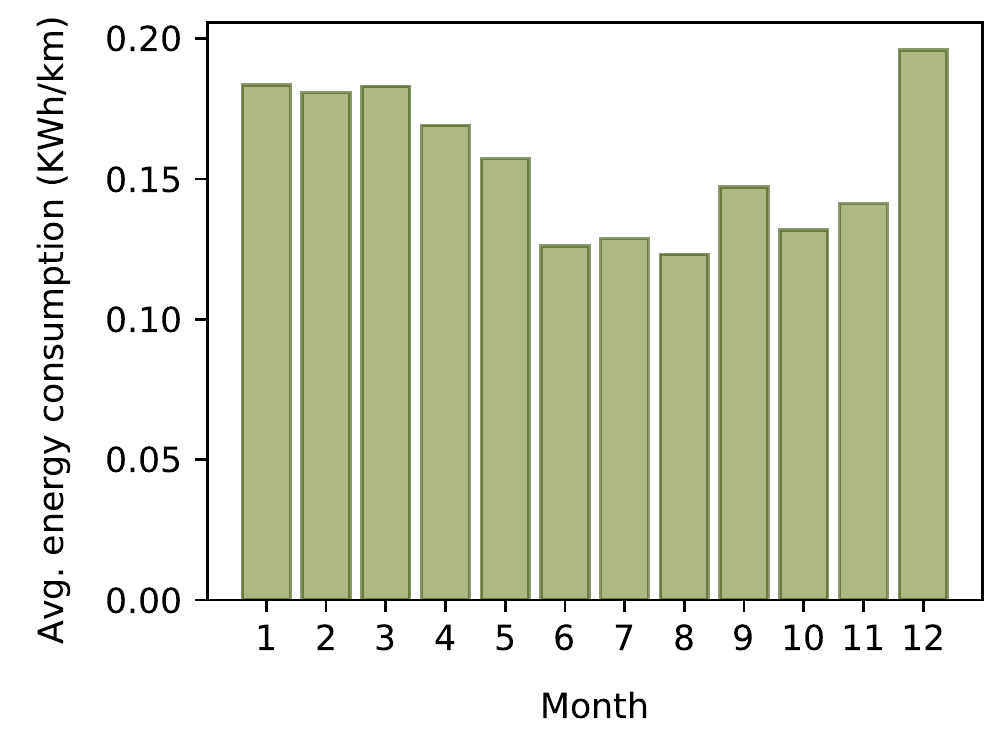}
    \caption{Temporal distribution.}
    \label{fig:fuelmonth}
  \end{subfigure}
  \begin{subfigure}{0.24\textwidth}
    \centering
    \includegraphics[width=\linewidth]{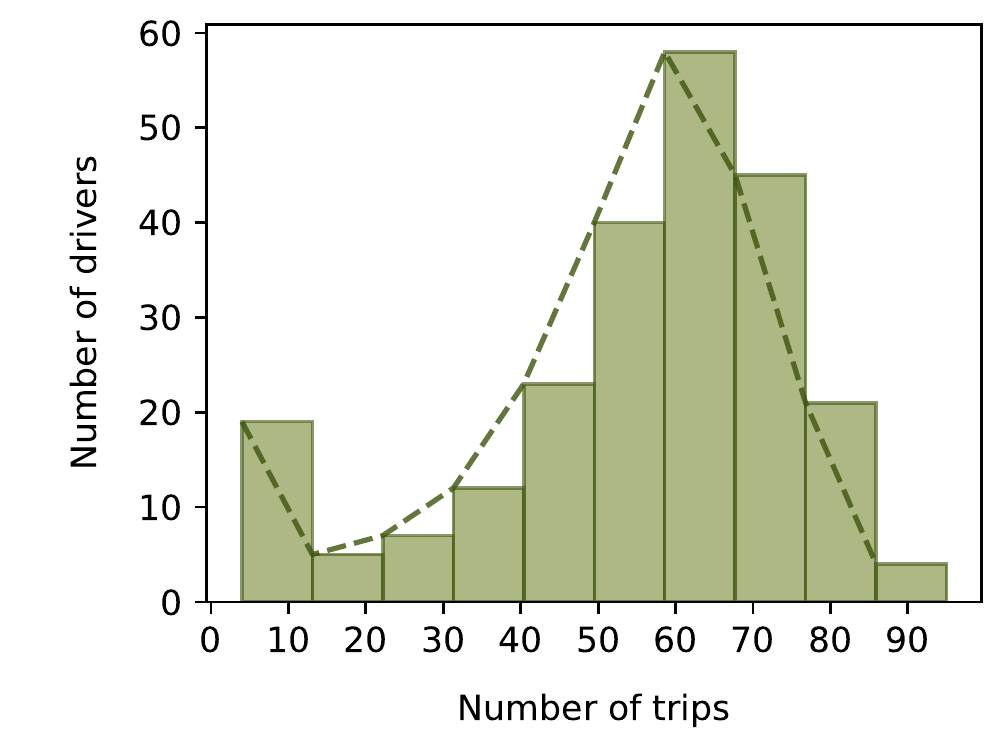}
    \caption{Trip frequency distribution.}
    \label{fig:tripnum}
  \end{subfigure}
  \caption{Distribution of the VED dataset: (a) the spatial distribution of vehicle energy consumption on each road segment. (b) the trip distance-energy distribution. (c) the temporal distribution of energy consumption in a year. (d) the driver-specific trip frequency distribution.}
  \label{fig}
\end{figure*}

\subsection{Data Description}
We study our problem on two real-world datasets. 
The first is a large-scale energy usage dataset of diverse vehicles in Ann Arbor, Michigan, USA, known as the vehicle energy dataset (VED\footnote{https://github.com/gsoh/VED}). Another is an electric taxi trajectory dataset (ETTD\footnote{http://guangwang.me/\#/data}), collected from Shenzhen, Guangdong, China. The statistics of two datasets are summarized in Table~\ref{tab:dataStats}.

The VED dataset \cite{oh2020vehicle} consists of vehicles' trajectories and the corresponding dynamic factors~(\ie energy, speed, \etc) collected by the Second On-Board Diagnostics (OBD-II) logger. 
The dataset is ranged from Nov 2017 to Nov 2018, covering various driving scenarios and weather conditions. 
In total, VED contains trajectories of approximately 130,864 kilometers. The fleet comprises 348 vehicles (230 Internal Combustion Engine Vehicles (ICEVs), 91 Hybrid Electric Vehicles (HEVs), 24 Plug-in Hybrid Electric Vehicles (HEVs), and 3 Electric Vehicles (EVs). 
We regard the fuel consumption of ICEVs and HEVs, and the electricity cost of PHEVs and EVs as the ground truth energy consumption~\cite{hua2022fine}.

The ETTD dataset \cite{wang2019experience} contains trajectories of 377 electric taxis with 1,155,654 GPS records and speed profiles collected on Oct 22, 2014. The driving conditions range from approximately all scenarios city-wide. The total distance is about 87,251 kilometers. We consider the required mechanical energy at the wheel as the ground truth energy consumption~\cite{xing2020energy} in the ETTD dataset.

\subsection{Data Preprocessing}
\subsubsection{Data Preparation}




We preprocess each dataset as follows. For each dataset, we split the raw trajectory into multiple trips if the driver has stopped for more than five minutes.
The route of each trip is extracted by a map-matching algorithm \cite{Yang2018FastMM}. We calculate the energy consumption on each GPS sample point by following \cite{hua2022fine}. 
Since the ETTD dataset only provides the speed and coordinate information, the VEC is calculated based on an estimation of the required mechanical energy at the wheel by following \cite{xing2020energy}. 
We aggregate the energy consumption of each GPS point as the ground truth of each trip.

\subsubsection{Data Anonymization}
The original datasets in our study did not include any identifiable driver information such as names, phone numbers, or any other personal details.
To further protect the sensitive information of each driver, we mask each driver with an anonymized identifier.
\eat{We carefully remove all potential individual data in the original dataset to make sure all records cannot be associated with sensitive personal information such as names and phone numbers.}

\subsection{Data Analysis}\label{ssec:dataanalysis}
To help understand the VEC distribution, we conduct primary data analysis on the VED dataset.
Overall, the energy consumption of each trip is influenced by various factors.
First, Figure \ref{fig:evenergy} plots the spatial distribution of energy consumption in Ann Arbor, where warmer color represents a higher energy consumption.
We observe the VEC varies in different road segments of the city, and vehicles cost more energy in the downtown area of the city than in the suburbs. 
Figure \ref{fig:distanceenergy} shows the positive correlation between trip length and energy consumption.
Meanwhile, Figure \ref{fig:fuelmonth} presents the varying energy consumption over the year, indicating a negative correlation between the energy consumption and temperature.
Finally, the trip frequency of each driver is reported in Figure \ref{fig:tripnum}.
We observe a two-peak distribution where the second-highest peak in the long-tail represents the large portion of drivers with only a few historical trips.
Such observation inspires us to develop a meta-optimization scheme to alleviate the cold-start problem.

\eat{
Figure \ref{fig:distanceenergy} shows the joint distribution of trip length and energy consumption. Figure \ref{fig:evenergy} shows the spatial distribution of energy consumption in Ann Arbor (warm colors represent higher VEC). Figure \ref{fig:fuelmonth} presents the average energy consumption in a year. Figure \ref{fig:tripnum} presents the trip number distribution of drivers. 
Although there is a near-linear relationship between the length of the trip and VEC, the spatial-temporal and personalized features are also significant factors. 
Firstly, we observe that vehicles cost more energy in the center area of the city than in the suburbs, indicating the VEC varies in different locations of the city. 
Moreover, drivers also need to open the air conditioning to maintain a comfortable temperature in the vehicle, which makes them consume much more energy in seasons like winter, indicating VEC also varies in different time periods. 
Meanwhile, we observe that some drivers only have a few trip records (\ie long-tail drivers). A globally well-trained model may fit these drivers terribly. It is necessary to design a method to handle the cold-start problem. Last but not least, VEC also varies with different driving behaviors. 
Figure \ref{fig:intro} presents the average energy consumption varying in individual drivers. Modeling driving preference would significantly help the model provide more accurate VEC estimations.
}
\section{THE PROPOSED METHOD}\label{sec:algo}


\begin{figure*}[!]
\centering
\includegraphics[width=1.0\textwidth]{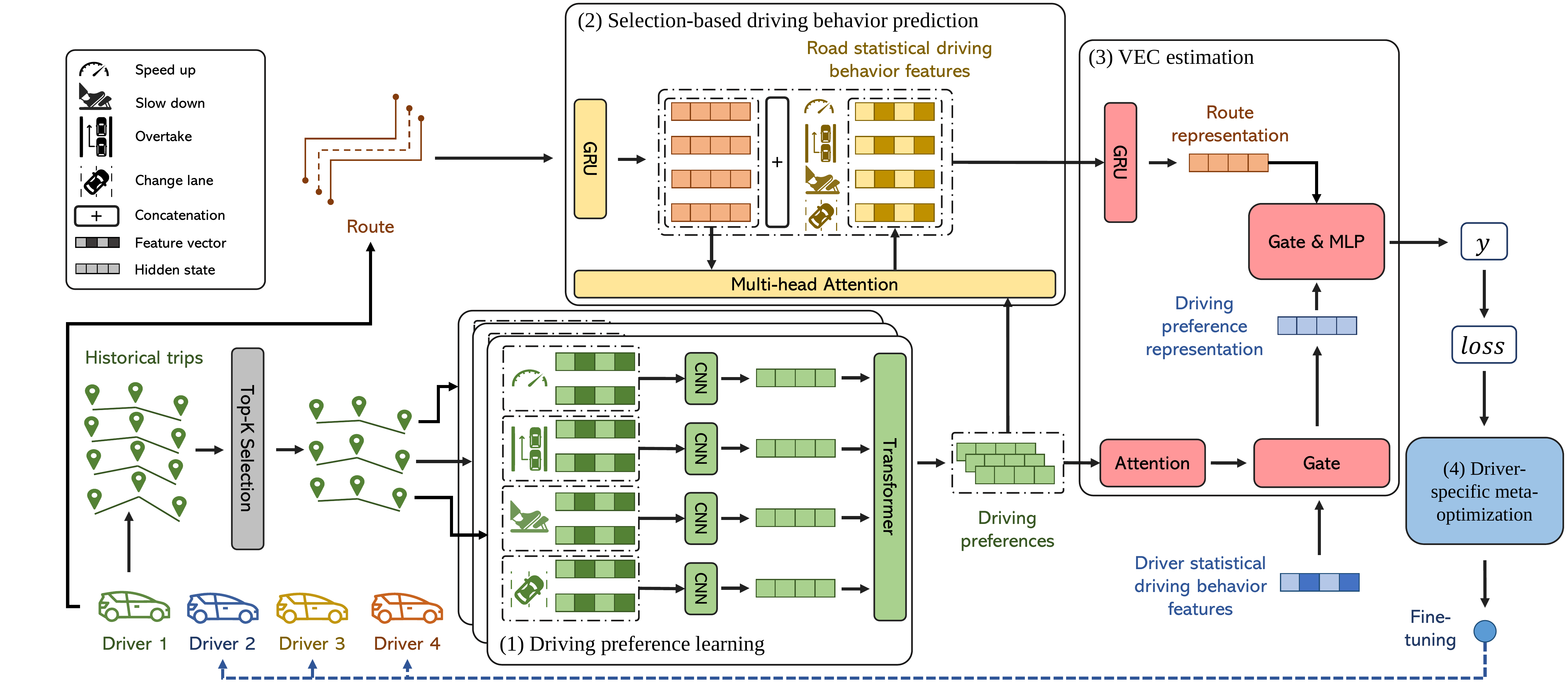}
\centering
\caption{The framework overview of \baby.}
\label{fig:archi}
\end{figure*}

The overall structure of our proposed \baby framework is illustrated in Figure~\ref{fig:archi}, which mainly consists of four components. (1) \textsl{Driving preference learning}: it performs spatiotemporal behavior learning to extract the driver's driving preferences from historical trips.
(2) \textsl{Selection-based driving behavior prediction}: it forecasts the driver's future driving behaviors on each road by jointly considering road conditions and driving preferences, providing the extra basis and supervision signals for VEC estimation. (3)\textsl{VEC estimation}: it predicts the VEC of the target trip. (4) \textsl{Driver-specific meta-optimization}: it learns the global parameter initialization and enables the model to fast adapt to each driver. We next present each component in detail.

\subsection{Feature Construction}
This section introduces the features we utilize based on the datasets mentioned above. Every feature vector will be projected into a $d$ dimensional representation via an embedding layer before feeding into the \baby model.

\subsubsection{The trip feature}
Trip features provide prior knowledge about the trip, including month, departure time, and route length.

\subsubsection{The statistical driving behavior feature}
The statistical driving behavior features provide macroscopic personalized information. We extract them from the driver's historical trips, including average speed, acceleration, VEC per hour, and VEC per kilometer. They are concatenated with the trip features and denoted as $\mathbf{x}^s$.
We also extract these features on each road of the target trip as the ground-truth labels for the selection-based driving behavior prediction module. The features on the road $e_i$ are denoted as $\mathbf{y}^e_i$.

\subsubsection{Vehicle state feature}
Vehicle state features describe the vehicle's current condition in the trajectory, including local time, speed, acceleration, VEC per hour, and VEC per kilometer.

\subsubsection{Road feature}
Road features describe the characteristics of the target road, which include the road type, the one-way indicator, the number of lanes, the allowed maximum speed, and the length of the road.

\subsection{Driving Preference Learning}\label{ssec:driPreExt}
Instead of measuring the driver's personalized information only by handcrafting statistical features, we propose to exploit fine-grained trajectory-level driving preferences hidden in historical trips. As demonstrated in Section \ref{ssec:dataanalysis}, VEC varies under different locations, time periods, and driver behaviors. We develop a spatiotemporal driving preference learning module to incorporate driving preference in an end-to-end way. 

\subsubsection{Distance-time encoding}
We first encode the location and time information in each trajectory in a normalized scale to ease spatiotemporal learning.
Inspired by the success of SeFT \cite{horn2020set} and Transformer \cite{vaswani2017attention} in preserving the location relation via position encoding,  we introduce distance-time encoding.
Specifically, we use sine and cosine functions of different frequencies to convert the 1-dimensional data axis into a multi-dimensional input, which is defined as follows:
\begin{align}
Enc^*_{2k}(a)&=sin(\frac{a}{\mathscr{a}^{2k/d}}),\\
Enc^*_{2k+1}(a)&=cos(\frac{a}{\mathscr{a}^{2k/d}}),
\end{align}
where $\mathscr{a}$ is the maximum value that is expected in the data, $k$ denotes the dimension, and $*$ can be an arbitrary encoding target. 
For distance encoding $Enc^{dist}(\cdot)$, $a$ stands for the distance the driver has traveled from the origin to the current location. 
For time-encoding $Enc^{time}(\cdot)$, $a$ represents the time elapsed since the trip started. Then, we attach two types of encodings into the vector of the vehicle state features:
\begin{equation}
\mathbf{x}^l_i = \mathbf{x}^l_i + Enc^{dist}(p_i) + Enc^{time}(t_i).
\end{equation}

\subsubsection{Driving behavior learning}
Then, we extract driving behaviors hidden in historical trajectories. 
We consider the driver's behaviors as a sequence of vehicle states, where each state $\mathbf{x}^l_i$ at a time stamp describes an instant condition of the vehicle (\eg current speed, acceleration, VEC per km, \etc).
We derive the latent driver behavior representation by analogous the complicated behavior as a semantic object in an image~\cite{dosovitskiy2020image}.
\eat{
Actually, a vehicle state $\mathbf{x}_i$ cannot be seen as a behavior. It is nothing more than a current condition (\eg current speed, acceleration, VEC per km, \etc). If we want to learn what he was doing at that time, we should consider a set of continuous states, which can be seen as behavior, representing actions like accelerating or braking. 
This process is similar to detecting objects in an image \cite{dosovitskiy2020image}. We replace pixels with the trip's vehicle states and deploy a CNN to extract their hidden features. }

Formally, we first split the vehicle state sequence of the historical trajectory $\mathbf{X}^l=[\mathbf{x}^l_1, \mathbf{x}^l_2, \cdots, \mathbf{x}^l_m]$ into multiple segments, then a convolutional neural network (CNN) is applied to embedding the driving behavior,
\begin{align}
\mathbf{Z}[i] = SELU(\mathbf{X}^l[q \cdot (i-1) + 1: q \cdot i] \odot \mathbf{F} + \mathbf{b}),
\end{align}
where $q < m$ is the number of states contained in a segment, $\mathbf{Z}[i]$ is the learned representation of the extracted behavior at step $i$, $\mathbf{Z}$ is the sequence of driving behaviors in each step, $\mathbf{F}\in \mathbb{R}^{q\times d}$ is the filter, $\odot$ denotes the convolution operation, $SELU$ is chosen as the activation function, and $\mathbf{b} \in \mathbb{R}^d$ is a learnable parameter.
We further employ a Transformer encoder to derive the unified driver preference representation $\mathbf{z}$,
\begin{align}
\mathbf{z} = TransformerEncoder(\mathbf{Z}).
\end{align}


\subsection{Selection-based Driving Behavior Prediction}\label{ssec: driBehPre}
Based on the driving preferences extracted above, we further infer the fine-grained trajectory, \ie the detailed driving behavior such as acceleration and speed on the target trip, to provide the additional basis and supervision signals for the VEC estimator.
Instead of predicting driving behaviors solely based on the road segment sequence, we propose to selectively reference similar routes of the target one to align the input~(\ie the feature vector) at training and inference time.
The selective approach also incorporates personalized information without introducing the significant computational overhead and extra noise.
\eat{
by simply taking the road embedding vectors as input, we propose referencing behaviors the driver took on historical trips and selecting the most probable ones the driver also would take according to the road conditions. On the one hand, it is a mechanism of behavior to behavior (B2B), which does not require the model to transform one domain to another (\ie road to behavior, R2B). On the other hand, this approach is driven by personalization, which forecasts the driver's driving behaviors based on what he (or she) would do in the past.
}

\subsubsection{Top-$K$ historical trip selection}\label{subsec:topk}
\eat{This study utilizes historical trips to extract the driver's driving preference. However, the number of them is relatively large, and some may contain lots of noisy information, we propose to select top-$K$ historical trips by keeping ones that covered as many of the same roads as the target trip, and the departure times also need to be close enough. To do this, we designed the route and departure time similarity functions to measure the score of each historical trip. Firstly, we can obtain the route similarity as the number of the same roads shared by the target route $R^c$ and the historical trip's route $R^o_i$ as:}

We propose to select top-$K$ historical trips to ease the driving behavior prediction.
Specifically, we construct two lightweight score functions to measure the distance between two trips.
The first distance function is based on the overlap of road segments,
\begin{equation}
score^{route}_{c, i} = \enspace \mid R_c \cap R_i \mid,
\end{equation}
where $\cap$ is the \textsl{intersection} operation. Meanwhile, we calculate the temporal similarity,
\begin{equation}
score^{time}_{c, i} = \enspace \mid s_c - s_i \mid,
\end{equation}
where $s_c$ and $s_i$ are the departure time of the target and the historical trip, respectively. Finally, we calculate the overall score of the historical trip,
\begin{align}
&norm(score^*_{c, i}) = \frac{score^*_{c, i} - Min(score^*_c)}{Max(score^*_c) - Min(score^*_c)},\\
&score_{c, i} = norm(score^{route}_{c, i}) - norm(score^{time}_{c, i}),
\end{align}
where $score^*_c$ denotes the similarity scores of the target trip compared with all historical trips of the driver. We select $K$ trips with the highest scores to extract the driver's driving preference. Note the reference trips may be less than $K$ for long-tail drivers with only a few historical trips.
Please refer to Appendix \ref{assc:topk} for the complete top-$K$ historical trip selection algorithm.

\subsubsection{Driving behavior prediction}
For driving behavior prediction, we first employ GRU~\cite{chung2014empirical}, a variant of the recurrent neural network, to encode contextual information of the road segment sequence in the target route:

\begin{align}
\mathbf{h}^e_i = GRU_1(\mathbf{h}^e_{i-1},\mathbf{x}^e_i;\bm{\Omega}_1),
\end{align}
where $\bm{\Omega}_1$ is parameters of GRU. With selected historical trips, then a multi-head attention module followed by a fully connected layer (FC) is applied to predict driving behaviors on each road segment:

\begin{align}
\label{equ:att_weight}
&\beta_{ij}=softmax\big(\frac{(\mathbf{W}_Q\mathbf{h}^e_i)^\top\mathbf{W}_K \mathbf{z}_j}{\sqrt{d}}\big),\\
&\mathbf{head} = \sum_{j=1}^K \beta_{ij} (\mathbf{W}_V\mathbf{z}_j),\\
&\hat{\mathbf{y}}^e_i = FC(\mathbf{head}_1 \parallel \dots \parallel\mathbf{head}_h),
\end{align}
where $\hat{\mathbf{y}}^e_i$ is the predicted statistical driving behavior features on the road $e_i$, $\mathbf{z}_j$ is the driving preference shown in the historical trip $j$, $\mathbf{W}_Q, \mathbf{W}_K, \mathbf{W}_V \in \mathbb{R}^{d \times d}$ are learnable parameters, $h$ is the number of heads, and $\parallel$ denotes the concatenation operation.

\subsection{VEC Estimation}
\eat{Now we have obtained features from five perspectives, which are the statistical driving behavior features (including trip features), driving preferences extracted in section \ref{ssec:driPreExt}, the target trip's driving behaviors predicted in section \ref{ssec: driBehPre}, and the route sequence.} 

Based on the five categories of features we obtained above, \ie the trip features, the statistical driving behavior features, the road features, the driving preferences extracted in section \ref{ssec:driPreExt}, and the target trip's driving behaviors predicted in section \ref{ssec: driBehPre}, we estimate the VEC of the target trip.

Firstly, we fuse all the driver's historical driving preference representations by an attention function as follows:
\begin{align}
\label{equ:hist_weight}
&\mu_i = softmax \big(\mathbf{W}_z(\mathbf{x}^s * \mathbf{z}_i) + \mathbf{b}\big),\\
&\mathbf{h}^z = \sum_{i=1}^K \mu_i \mathbf{z}_i,
\end{align}
where $\mathbf{h}^z$ is the representation of driving preference extracted from top-$K$ historical trips, $\mathbf{W}_z \in \mathbb{R}^{d \times d}$ and $\mathbf{b} \in \mathbb{R}^d$ are learnable parameters, and $*$ denotes element-wise multiplication.

For the road segment sequence, we concatenate each road segment's embedding vector with its corresponding predicted statistical driving behavior features as $\hat{\mathbf{x}}^e_i = [\mathbf{x}^e_i \parallel \hat{\mathbf{y}}^e_i]$. Then, we adopt another GRU to encode the updated contextual relations among roads:
\begin{align}\label{equ:roadembedding}
\hat{\mathbf{h}}^e_i = GRU_2(\hat{\mathbf{h}}^e_{i-1},\hat{\mathbf{x}}^e_i;\bm{\Omega}_2).
\end{align}
We take the last hidden state $\hat{\mathbf{h}}^e_n$ from \eqref{equ:roadembedding} as the representation of the target route.

Afterward, we propose a gating mechanism to fuse the representation of driving preference, the statistical driving behavior features, and the target route,
\begin{align}
&\mathbf{h}= (\mathbf{h}^z \star \mathbf{x}^s) \star \hat{\mathbf{h}}^e_n,
\end{align}
where $\mathbf{h}$ is the representation of the target route, and $\star$ denotes the parameterized gating mechanism.

Finally, we estimate the VEC of the target route by feeding $\mathbf{h}$ into a multi-layer perception (MLP), then multiply it with the vehicle type embedding $\mathbf{W}_{tp} \in \mathbb{R}^d$:

\begin{align}
\mathbf{h} &= MLP(\mathbf{h}),\\
\hat{y} &= \mathbf{W}_{tp}^\top \mathbf{h},
\end{align}
where $tp \in \left\{ICEV, HEV, PHEV, EV\right\}$ indicates the type of the vehicle.

\eat{Finally, we estimate the VEC of the target route based on a multi-layer perception,
\begin{align}
\hat{y} &= MLP(\mathbf{h}).
\end{align}}

\subsection{Driver-specific Meta-optimization}
A well-trained global model may fail to estimate the VEC for long-tail drivers with insufficient historical trips.
Inspired by the success of Model-Agnostic Meta-Learning (MAML) \cite{finn2017model} in handling few-shot problems, we propose to learn a meta-optimized universal parameter initialization that can fast adapt to all drivers.
Then we fine-tune private models for each driver based on the model initialized by meta-training.

Formally, we apply MAE as our loss function for both driving behavior prediction and VEC estimation.
\begin{align}
&\mathcal{L}_{beh}(\hat{\mathbf{y}}^e) = \frac{1}{n}\sum^n_{i=1} \sum^f_{j=1} \mid \mathbf{y}^e_i[j] - \hat{\mathbf{y}}^e_i[j] \mid,\\
&\mathcal{L}_{EC}(\hat{y}) = \enspace \mid y - \hat{y} \mid,\\
&\mathcal{L} = \mathcal{L}_{beh}(\hat{\mathbf{y}}^e) + \mathcal{L}_{EC}(\hat{y}),
\end{align}
where $j$ and $f$ denote the index and the number of statistical driving behavior features, respectively.

Then, we regard VEC estimation for an individual driver as an independent task. The drivers' datasets can be denoted as $\{(D_s^u, D_q^u)\}^U_{u=1}$, where $U$ is the number of drivers, $D_s^u$, $D_q^u$ denotes the support set, and the query set of driver $u$, respectively.

In $i$-th epoch, we run every dataset on MAML to learn the globally adaptive model parameters. In every meta-training step, we perform fast adaptation on the support set $D_s^u$, calculate the loss $\mathcal{L}^i_{D^u_s}(f_\theta)$, and update parameters as follow:

\begin{align}
\theta^\prime \leftarrow \theta - \eta \nabla_\theta \mathcal{L}^i_{D_s^u}(f_\theta),
\end{align}
where $\theta^\prime$ are updated parameters, $\eta$ is the inner loop learning rate, and $\nabla_\theta \mathcal{L}^i_{D^u_s}(f_\theta)$ denotes the gradient of loss on the support set. Then, we evaluate the model by running the query set $D_q^u$ and calculate the loss $\mathcal{L}^i_{D^u_q}(f_{\theta^\prime})$. At the end of each epoch, we apply bi-level optimization to update model parameters as:

\begin{align}
\theta \leftarrow \theta - \gamma \nabla_\theta \sum^U_{u=1} \mathcal{L}^i_{D_q^u}(f_{\theta^\prime}),
\end{align}
where $\gamma$ is the outer loop learning rate. We run several meta-training epochs until it performs well on the validation dataset. 

Afterward, we merge the support and query set as $D^u$, then fine-tune the initialized parameters on driver $u$. In the end, we would obtain $U$ sets of parameters, and each of them can be well adapted to its corresponding driver. Please refer to Appendix \ref{assc:meta} for the complete meta-optimization algorithm.
\section{EXPERIMENTS}

\subsection{Experimental Setup}
In this section, we introduce the metrics of our experiments, the baseline models we compared, and the implementation details. Moreover, please refer to \ref{subsec: prototype} for the prototype system design.

\subsubsection{Metrics}
We adopt three widely used evaluation metrics in regression tasks: Mean Square Error (MSE), Mean Absolute Error (MAE), and Mean Absolute Percentage Error (MAPE).

\subsubsection{Implementation Details}
The embedding size of features is set as 20. The top-$K$ value is 5. We take the filter size $q=4$ for CNN and the number of heads $h=4$ for the Transformer encoder and the multi-head attention module. The hidden size of GRU and MLP modules is set as 20 and 40, respectively. We utilize Adam optimizer with the inner loop learning rate $6\times 10^{-4}$ and $3\times 10^{-4}$, the outer loop step size $6\times 10^{-3}$ and $3\times 10^{-3}$, and the fine-tuning learning rate $6\times 10^{-4}$ and $10^{-4}$ for the VED and ETTD datasets. The L2 penalty is set to $10^{-5}$. For each dataset, we randomly select each driver's 10\% and 20\% of records as the validation and test sets, respectively. The rest is left for training. The support set is split from the training set of the driver with a ratio of $10\%$, and the rest is left as the query set.

\subsubsection{Baselines}
We compare \baby\footnote{Source code is available at \url{https://github.com/usail-hkust/Meta-Pec}} with numerical and data-driven methods. The following ten baselines are compared.\\
\textbf{Average}~\cite{de2017data} estimates the energy consumption by multiplying the VEC per kilometer by the length of the trip.\\
\textbf{MLR} is known as the multiple linear regression model.~\cite{de2017data} utilizes real-world measured driving data and domain knowledge to construct a particular MLR model for VEC estimation, and predicts the speed profile based on a NN model.\\
\textbf{XGBoost} is a well-known gradient boosting model. ~\cite{chen2022prediction} utilizes XGBoost as the state-of-the-art model.\\
\textbf{DNN}~\cite{de2017data, roy2022reliable, chen2021data} is a classic NN model and widely used by multiple researchers in VEC estimation tasks.\\
\textbf{LSTM}~\cite{chen2021data, moawad2021deep} is a typical recurrent neural network widely utilized to handle the route sequence data and make road-level VEC estimations.\\
\textbf{Transformer}~\cite{vaswani2017attention} is a well-known self-attention-based sequential model.\\
\textbf{LDFeRR}~\cite{liu2021ldferr} utilizes an attention-based GRU to estimate the road-level VEC.\\
\textbf{Enc-Dec}~\cite{moawad2021deep} utilizes an encoder-decoder structure to estimate the road-level VEC.\\
\textbf{PLd-FeRR}~\cite{wang2022personalized} identifies the features indicating the driving preference, and a Transformer-based model is deployed for VEC estimation.\\
\textbf{Meta-TTE}~\cite{wang2022fine} is a meta-learning-based model utilized for the estimated time of arrival (ETA) prediction tasks. We modify its output from travel time into the energy consumption of the target trip.\\

\begin{small}
\begin{table}[t]
\setlength\tabcolsep{4pt}
\centering
\caption{Overall performance of \baby and all baselines on two datasets. The best and second-best results are highlighted in boldface and underlined, respectively.}
\label{tab:overall}
\begin{tabular}{ccccccc}
\toprule
\multirow{2}{*}{Model}&
\multicolumn{3}{c}{VED}&\multicolumn{3}{c}{ETTD}\cr
\cmidrule(lr){2-4} \cmidrule(lr){5-7}
                       & MSE    & MAE    & MAPE    & MSE    & MAE    & MAPE    \\
\midrule
Average                   & 0.4142 & 0.1569 & 62.52\% & 0.8245 & 0.5279 & 43.27\% \\
MLR                    & 0.0551 & 0.0529 & 48.12\% & 0.7627 & 0.5014 & 38.95\% \\
XGBoost                & 0.0446 & 0.0416 & 28.05\% & 0.7531 & \underline{0.4989} & 42.31\% \\
\midrule
DNN                    & 0.0397 & 0.0427 & 26.55\% & 0.7956 & 0.5187 & 38.38\% \\
LSTM                    & 0.0393 & 0.0421 & 27.56\% & 0.7646 & 0.5045 & 39.36\% \\
Transformer            & 0.0382 & 0.0398 & 24.50\% & 0.7669 & 0.5068 & 39.30\% \\
\midrule
LDFeRR                 & 0.0470 & 0.0425 & 28.16\% & 0.7483 & 0.5085 & 38.76\% \\
Enc-Dec           & \underline{0.0378} & 0.0398 & 25.55\% & 0.7496 & 0.5071 & 37.81\% \\
PLd-FeRR              & 0.0405 & \underline{0.0395} & \underline{24.37\%} & 0.7519 & 0.504  & \underline{37.66\%} \\
\midrule
Meta-TTE               & 0.0458 & 0.0463 & 40.72\% & \underline{0.7416} & 0.5139 & 40.12\% \\
\midrule
\baby                   & \textbf{0.0350} & \textbf{0.0349} & \textbf{22.82\%} & \textbf{0.7306} & \textbf{0.4898} & \textbf{37.10\%} \\
\bottomrule
\end{tabular}
\end{table}
\end{small}
\vspace{-0.54cm}

\subsection{Overall Results}
Table \ref{tab:overall} shows the experimental results of the ten baselines and our proposed method on two real-world datasets. \baby outperforms all the baseline models and achieves state-of-the-art. In detail, almost all classic methods (Average, MLR, XGBoost) perform worse than deep learning models since they only consider the trip's and driver's statistical information. They estimate energy consumption from a macroscopic view. 
Well-designed VEC estimation-oriented models (LDFeRR, Enc-Dec, PLd-FeRR) have the best performance among most baseline models, indicating more information is captured by the model as we consider more driver-specific features. Meta-TTE is a model for ETA prediction, which performs worse on the VED since this dataset is constructed based on real vehicle energy usage. In contrast, VECs of the ETTD dataset are calculated based on mechanical energy, indicating a gap existing between the ETA and VEC prediction tasks. 
Note Meta-TTE is designed to provide accurate travel time estimates even when there are changes in traffic conditions or road networks, while our model leverages meta-learning to acquire globally shared knowledge from various drivers and enable rapid adaptation to insufficient data scenarios.

Further looking into the results, \baby significantly surpasses all deep-learning-based models by ($7.51\%$, $11.66\%$, $6.37\%$) on the VED dataset, indicating our proposed method can make more appropriate driver-specific VEC estimations. The reasons are three-fold. First, unlike other baseline models which only model personalization features by handcrafting statistical features, we extract driving preferences from the historical trajectories at the trajectory-level. Second, the driving behavior prediction module also helps in providing extra estimation basis and supervision signals. More in-detailed analysis of this module will be provided in Section \ref{sssec:abprediction}. 
Finally, different from methods that learn a unified model for all users, we utilize a meta-optimization strategy to perform fast adaptation on long-tail drivers, which will be studied in Section \ref{ssec:coldstart}.

\subsection{Ablation Study}
\begin{figure}[!]
\centering
\includegraphics[width=\columnwidth]{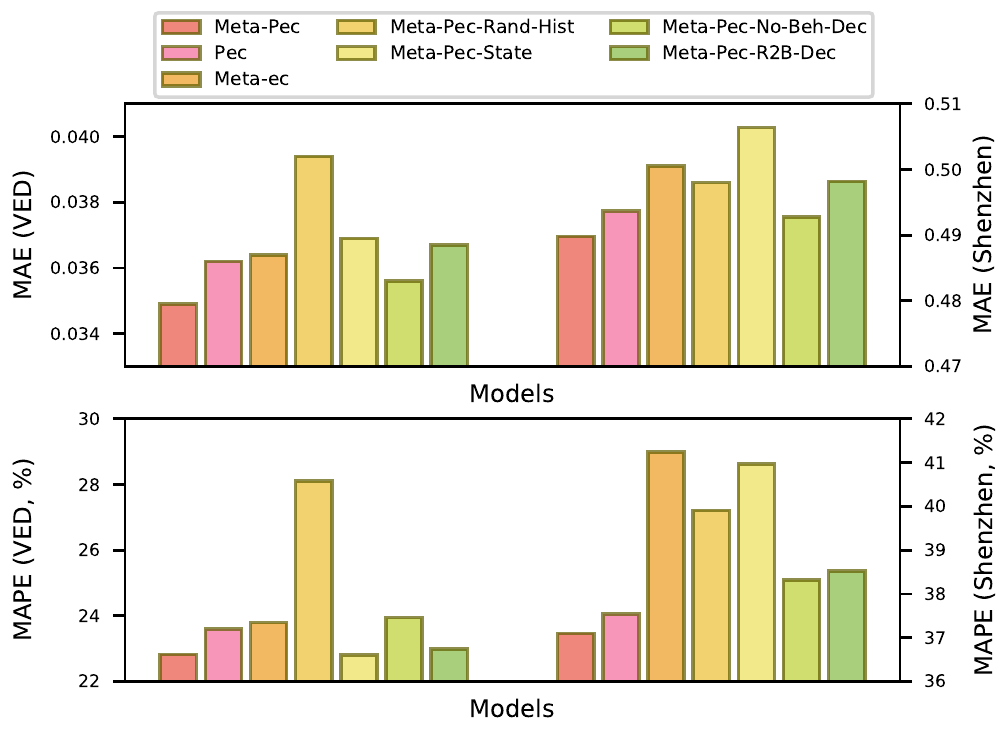}
\centering
\caption{Ablation tests of the model on two datasets.}
\label{fig:ablation}
\end{figure}

In order to verify the effectiveness of each module, we conduct ablation studies on six variants of our proposed \baby, including (1)~\textsc{Pec}: the base model, which does not utilize the meta-optimization strategy, (2)~\textsc{Meta-ec}: the model does not use any personalization module (\ie the driving preference learning and driving behavior prediction module), (3)~\textsc{Meta-Pec-Rand-Hist}: the model randomly selects $K$ historical trips to extract the driver's preference, (4)~\textsc{Meta-Pec-State}: the model learns the driving preference by modeling vehicle state sequence rather than driver behaviors, (5)~\textsc{Meta-Pec-No-Beh-Dec}: the model that does not predict driving behaviors on each target road, and (6)~\textsc{Meta-Pec-R2B-Dec}: the model that predicts driving behaviors only based on the road features rather than jointly considering driving preferences and the road conditions. The comparison results among all variants are shown in Figure \ref{fig:ablation}.

\subsubsection{The effectiveness of the meta-optimization module (\textsc{Pec})}
Although \textsc{Pec} and \baby have similar performance, \baby is better (3.74\% at most on the VED dataset) due to its meta-optimization module, indicating the model has learned the globally shared knowledge and is able to adapt quickly to each driver's particular preference and provide accurate VEC estimations for long-tail drivers.

\subsubsection{The effectiveness of personalization modules (\textsc{Meta-ec})}
We designed two modules for personalized VEC estimation, including the driving preference learning module and the selection-based driving behavior prediction. After we exclude these two components and only leave statistical information as personalized features, it performs badly on the two datasets (drops 11.16\% at most on the ETTD dataset), indicating the handcrafting statistical data is coarse-grained and helps little to the accurate personalized estimations.

\subsubsection{The effectiveness of the top-$K$ historical trip selection (\textsc{Meta-Pec-Rand-Hist})}
After we replace the top-$K$ historical trip selection strategy by random picking, this variant model performs much worse on two datasets (drops 23.19\% at most on the VED dataset), representing the irrelevant trips provide little information about how the driver will drive on the target route.

\subsubsection{The effectiveness of modeling driver behaviors (\textsc{Meta-Pec-State})}
\textsc{Meta-Pec-State} does not deploy CNN to extract behaviors but only uses a transformer module to model the vehicle state sequence. It performs terribly on the two datasets (drops 10.43\% at most on the ETTD dataset), indicating it is reasonable to encode driver behaviors rather than directly model vehicle states on the trajectory.

\subsubsection{The effectiveness of driving behavior prediction (\textsc{Meta-Pec-No-Beh-Dec})}\label{sssec:abprediction}
The driving behavior prediction module can offer more VEC estimation basis for VEC estimations and extra supervision signals as a joint learning module. As we exclude the driving behavior prediction module, the performance drops on the two datasets (4.91\% at most on the VED dataset).

\subsubsection{The effectiveness of the Behavior to Behavior (B2B) prediction (\textsc{Meta-Pec-R2B})}
Previous studies like \cite{moawad2021deep} predict road-level driver energy consumption only based on road features, which require the model to transfer the road domain into the energy usage behavior domain (R2B). It is a more complex manner and does not consider any personalization information. After we switch B2B into R2B prediction, the performance drops on the two datasets (5.17\% at most on the ETTD dataset).

\begin{small}
\begin{table}[t]
  \centering
  \caption{Performances on long-tail drivers}
  \label{tab:coldstart}
    \begin{tabular}{ccccccc}
    \toprule
    \multirow{2}{*}{Model}&
    \multicolumn{3}{c}{VED}&\multicolumn{3}{c}{ETTD}\cr
    \cmidrule(lr){2-4} \cmidrule(lr){5-7}
    & MSE & MAE & MAPE & MSE & MAE & MAPE \cr
    \midrule
    \textsc{Pec}  & 0.0452 & 0.0686 & 22.85\% & 2.9598 & 1.0203 &70.41\%\cr
    \baby  & \textbf{0.0316} & \textbf{0.0587} & \textbf{21.86}\% & \textbf{2.8950} & \textbf{0.8848} & \textbf{50.25}\%\cr
    \bottomrule
    \end{tabular}
\end{table}
\end{small}

\subsection{Effectiveness on Long-tail Drivers}\label{ssec:coldstart}
We consider drivers who have less than ten training samples as long-tail drivers, which account for 14.65\% and 8.12\% in the VED and ETTD datasets, respectively. To verify the effectiveness of our proposed meta-optimization module, we further compare the performance of \baby and \textsc{Pec} (\ie the model does not utilize the meta-optimization module) on long-tail drivers. Table \ref{tab:coldstart} presents the results on two datasets with and without the meta-optimization module. We observe that \baby significantly surpasses \textsc{Pec} on long-tail drivers (30.08\% and 28.64\% at most on the VED and ETTD dataset, respectively), indicating our proposed model is more robust in handling drivers with limited instances and achieving personalized VEC estimation for various drivers.

\subsection{Parameter Sensitivity}
We conduct experiments on the two datasets to study the impacts of the following hyper-parameters in \baby. 
$K$ is the number of the most similar historical trips we extracted.
$q$ is the number of continuous vehicle states in a behavior segment.
States contained in a segment are considered a driving behavior in the trip trajectory (\eg accelerating, braking, etc).


\begin{figure}[t]
  \begin{subfigure}{0.49\columnwidth}
    \centering
    \includegraphics[width=\linewidth]{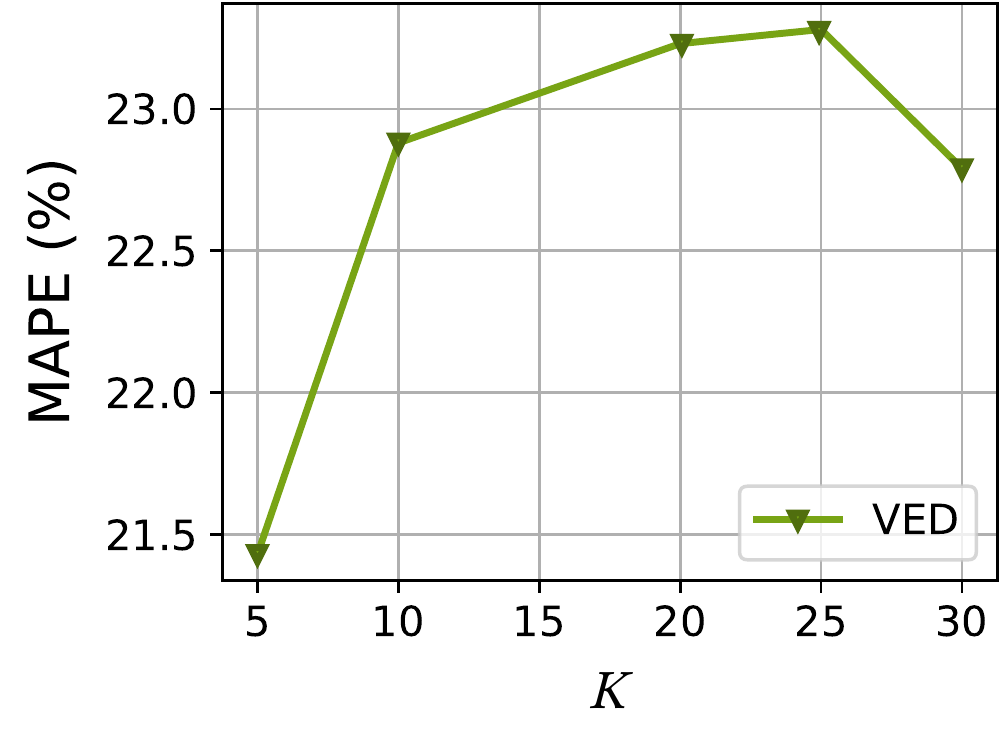}
    \caption{MAPE ($K$, VED).}
    \label{fig:sensiKVED}
  \end{subfigure}
  \begin{subfigure}{0.49\columnwidth}
    \centering
    \includegraphics[width=\linewidth]{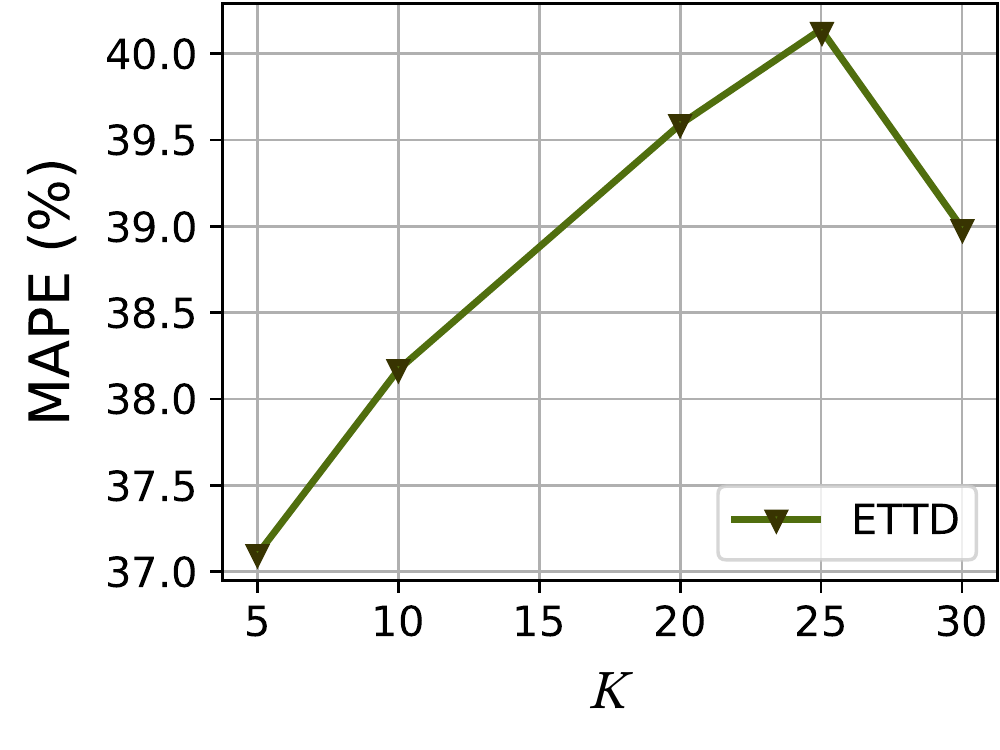}
    \caption{MAPE ($K$, ETTD).}
    \label{fig:sensiKETTD}
  \end{subfigure}
  \begin{subfigure}{0.49\columnwidth}
    \centering
    \includegraphics[width=\linewidth]{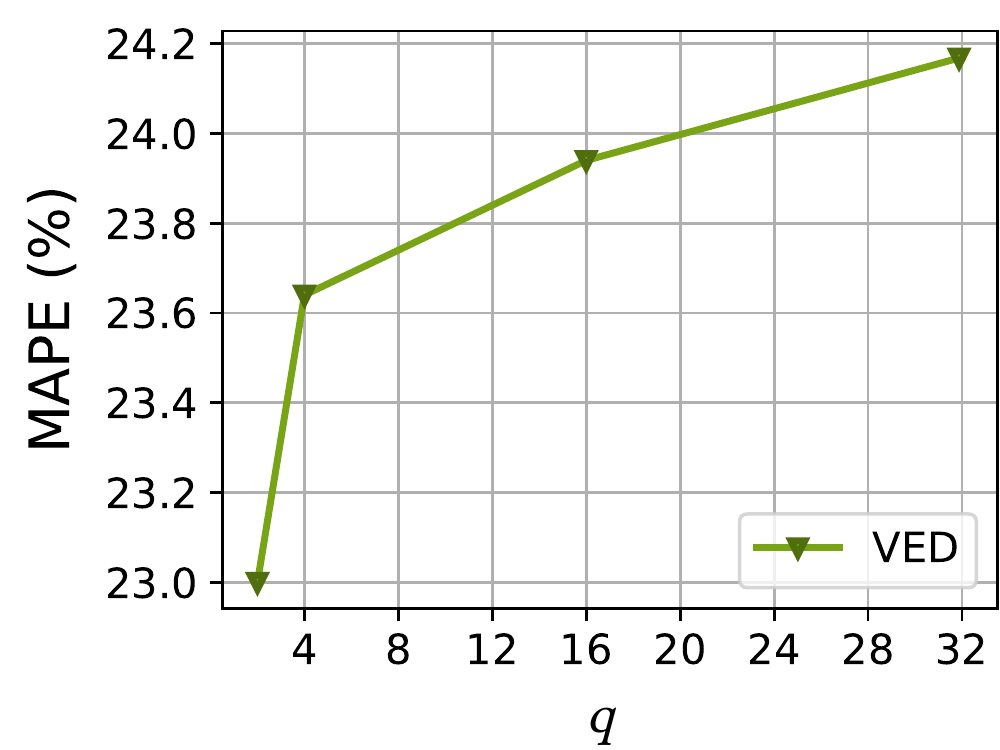}
    \caption{MAPE ($q$, VED).}
    \label{fig:sensiqVED}
  \end{subfigure}
  \begin{subfigure}{0.49\columnwidth}
    \centering
    \includegraphics[width=\linewidth]{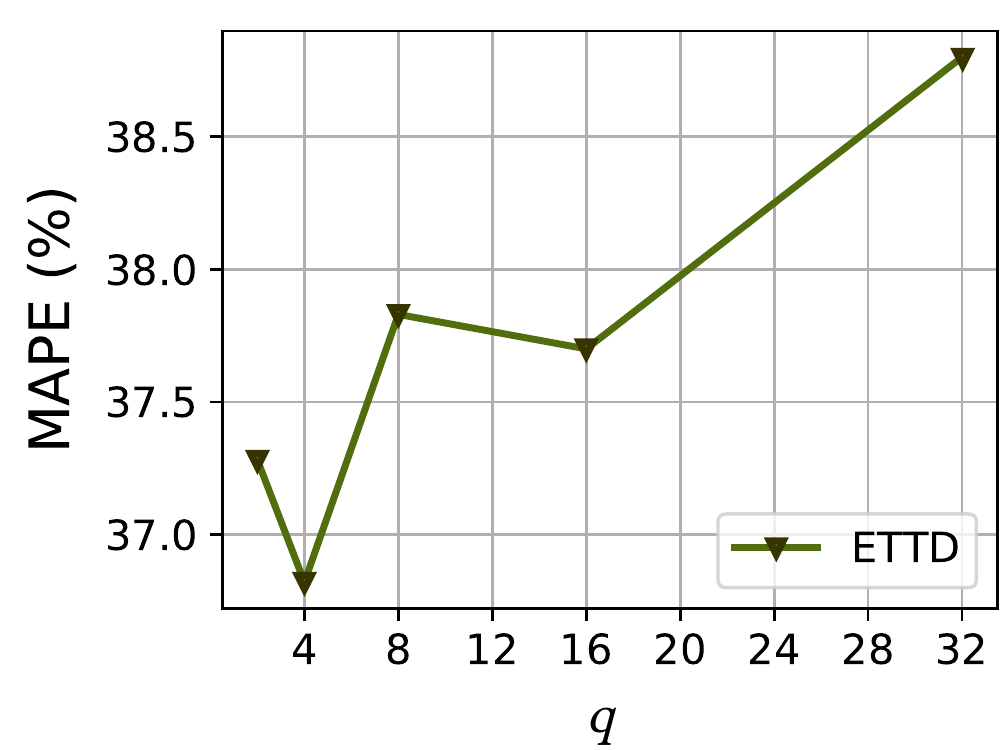}
    \caption{MAPE ($q$, ETTD).}
    \label{fig:sensiqETTD}
  \end{subfigure}
  \caption{Parameter sensitivity tests of the model on two datasets.}
  \label{fig:sameroadpred}
\end{figure}

\begin{figure}[t]
  \begin{subfigure}{0.48\columnwidth}
    \centering
    \includegraphics[width=0.8\linewidth]{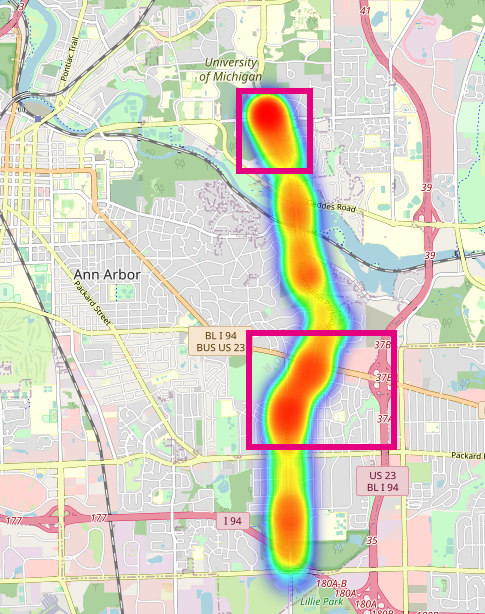}
    \caption{The ground truth VEC.}
    \label{fig:groundtruth}
  \end{subfigure}
  \begin{subfigure}{0.48\columnwidth}
    \centering
    \includegraphics[width=0.8\linewidth]{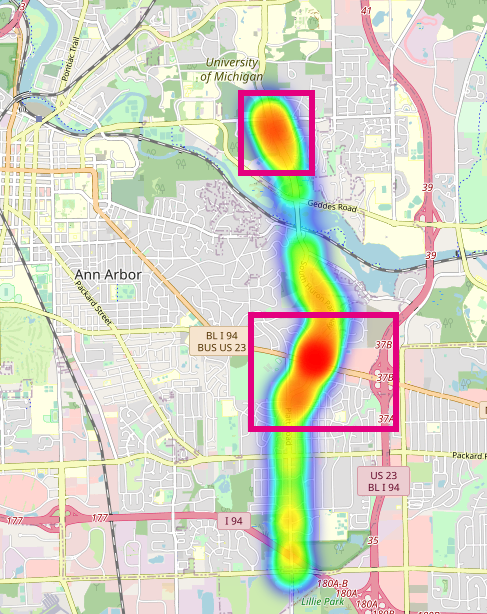}
    \caption{The predicted VEC.}
    \label{fig:pred}
  \end{subfigure}
  \caption{The spatial distribution of the ground truth and predicted VEC.}
  \label{fig:sameroadpred}
\end{figure}

\begin{table}[t]
    \caption{The importance of each top-$K$ historical trips. "\# of same roads" indicates the number of road segments shared by the historical and the target route.}
    \label{tab:weight}
    \begin{tabular}{cccc}
      \toprule
      \textbf{Rank}&\textbf{\# of same roads}&\textbf{Importance}\\
      \midrule
      \# 1 & 31 & 0.2689 \\
      \# 2  & 29 & 0.2626 \\
      \# 3 & 29 & 0.2637 \\
      \# 4 & 25 & 0.1074 \\
      \# 5 & 24 & 0.0974 \\
    \bottomrule
  \end{tabular}
\end{table}

\begin{figure}[t]
\centering
\includegraphics[width=1.0\columnwidth]{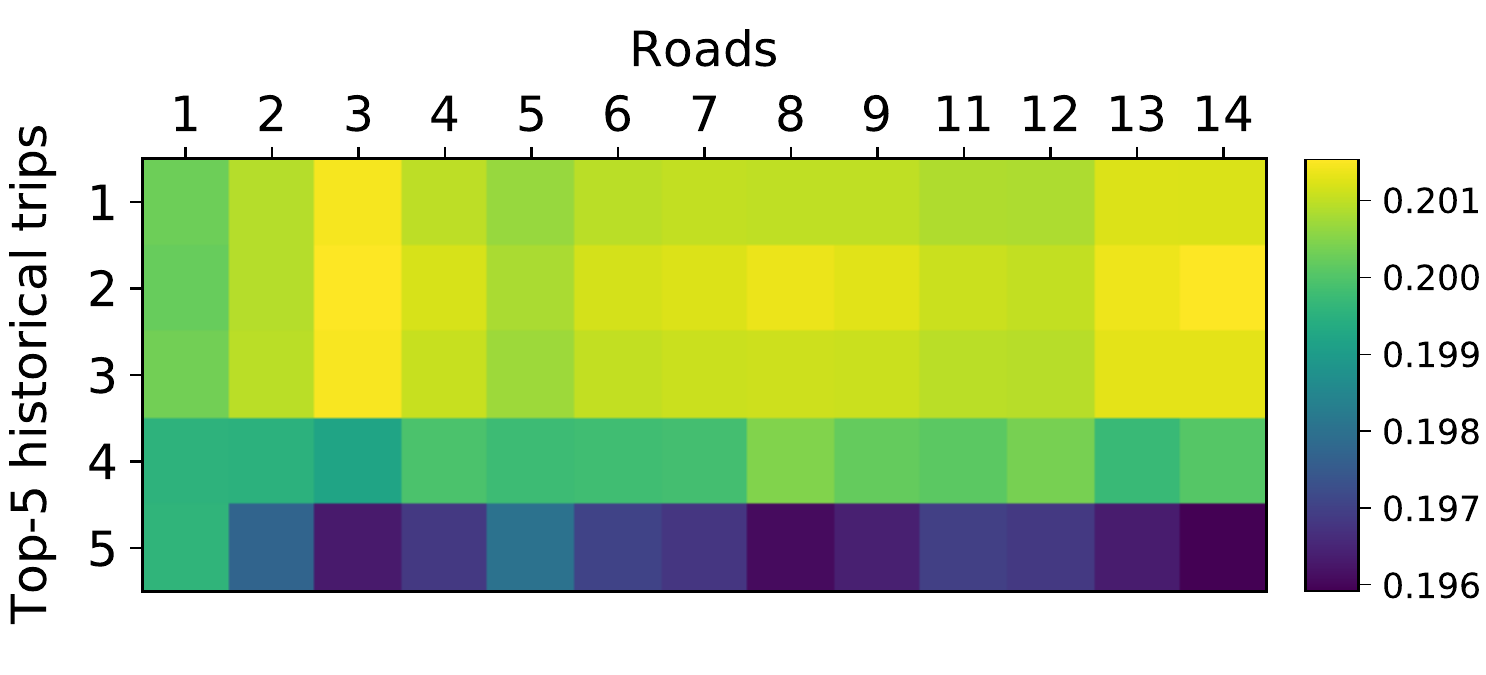}
\centering
\caption{The learned multi-head attention weights.}
\label{fig:multiheadattheatmap}
\end{figure}

Figure~\ref{fig:sensiKVED} and Figure~\ref{fig:sensiKETTD} show the results with varying $K$. It is clear that the model's performance drops as we sample more historical trips, indicating dissimilar trips contain noisy information and are less helpful to the driver's driving preference. Keeping the most relevant trips can help the model focus on what is truly important. \baby achieves the best performance when setting $K=5$ on the two datasets.

Figure~\ref{fig:sensiqVED} and Figure~\ref{fig:sensiqETTD} report the results of varying $q$. We observe that the performance drops when more states are included in a behavior segment, which is perhaps because the CNN needs to handle more behaviors simultaneously. As a result, the model performance decreased as the task became difficult. We set the $q$ value as 2 and 4 on the VED and ETTD datasets, respectively.

\subsection{Case Study}
In this section, we conduct a case study to further validate the VEC estimation performance of our proposed \baby. We take an example of the VED dataset, which achieves absolute error (AE) at $5 \times 10^{-4}$ and absolute percentage error (APE) at 0.56\%. 

\subsubsection{Spatial visualization}
We first analyze the spatial distribution of the estimated VEC.
Figure \ref{fig:groundtruth} shows the route and the ground truth energy usage of the example route. Figure \ref{fig:pred} shows the energy consumption on each road segment predicted by the selection-based driving behavior prediction module. 
As can be seen, our model properly estimated the high energy consumption area~(in red rectangles) indicating \baby successfully captures the driver's preference and road conditions for accurate VEC estimations.

\subsubsection{Module-level analysis}
We report the learned importance of historical trips in Table \ref{tab:weight}.
As can be seen, the importance of historical trips decreases gradually, indicating \baby owns the ability to discriminate the most critical historical trips that can provide helpful information to the VEC estimation. 
Besides, we calculate the multi-head attention weights of each historical trip defined by Equation \ref{equ:att_weight}. As illustrated in Figure \ref{fig:multiheadattheatmap}, the learned attention weights of top-3 trips are higher than the rest two, indicating that our driving behavior prediction module has also learned how to select the most useful information.
\section{RELATED WORK}

\subsection{Vehicle Energy Consumption Estimation}
Previous studies on vehicle energy consumption estimation can be mainly categorized into numerical approaches \cite{de2015energy, al2021driving, ojeda2017fuel, ding2017greenplanner} and data-driven approaches \cite{de2017data, chen2021data, liu2021ldferr, elmi2021deepfec, petkevicius2021probabilistic, hua2022fine, moawad2021deep, wang2022personalized}. 
Based on the vehicle dynamics equation as the underlying physical model, Cauwer \etal \cite{de2015energy} proposed using multiple linear regression (MLR) models to identify correlations between the kinematic parameters of the vehicle and VEC. Similarly, Al-Wreikat \etal \cite{al2021driving} proposed to evaluate the driving behavior, distance, temperature, traffic, and road grade effects on the VEC of an electric vehicle and the result shows the driver's behaviors have significant influences on VEC.
Ojeda \etal \cite{ojeda2017fuel} considered using a real physical vehicle model for the speed and fuel consumption prediction. Ding \etal \cite{ding2017greenplanner} proposed to evaluate VEC by measuring the power generated by fuel combustion.
Recently, machine learning based methods have been studied for vehicle energy consumption estimation.
Cauwer \etal \cite{de2017data} developed a data-driven method utilizing real-world measured driving data and domain knowledge to construct MLR models for VEC estimation. 
Chen \etal \cite{chen2021data} proposed to use of long short-term memory (LSTM) and artificial neural network (ANN) models to estimate the energy consumption of electric buses. 
Liu \etal \cite{liu2021ldferr} utilized an attention-based GRU to estimate the road-level VEC. 
DeepFEC \cite{elmi2021deepfec} proposed a deep-learning-based model to forecast energy consumption on every road in a city based on real traffic conditions. 
Hua \etal \cite{hua2022fine} developed a transfer learning model for electric vehicle energy consumption estimation based on insufficient electric vehicles and ragged driving trajectories.
PLd-FeRR \cite{wang2022personalized} employs the Transformer for VEC estimation, with a consideration of the driver's driving preference through handcrafted personalized features.

\subsection{Spatial-temporal Data Mining}

Our study is closely related to spatiotemporal data mining~\cite{wang2018learning, huang2022dueta, hong2020heteta, DBLP:conf/kdd/XuZZLZCX16, rao2022graph, xu2022metaptp, zhang2021intelligent, zhang2022multi, fliupractical2022, zhou2019collaborative}.
With the propensity of GPS devices, spatial-temporal data mining has been extensively studied in various applications.
To name a few, Estimated time of arrival (ETA) prediction is a classic task that aims to estimate the travel time with a given origin, destination, and departure time. WDR \cite{wang2018learning} proposed to combine Wide-Deep Learning with an LSTM module for ETA. It is worth to mention that WDR also considers some personalized features (\eg driver profile, rider profile, vehicle profile, \etc) to improve the performance. 
HetETA \cite{hong2020heteta} proposes to transform the road map into a heterogeneous graph and introduce a vehicle-trajectories-based network to consider traffic behavior patterns jointly. 
Huang \etal \cite{huang2022dueta} argue modeling traffic congestion is important for accurate ETA prediction, and they constructed a congestion-sensitive graph and a route-aware graph transformer to learn the long-distance congestion correlations.
Driving is a complex activity influenced by multiple factors. Analyzing driving behavior helps to assess driver performance, including safety and energy efficiency, leading to enhancements in transportation systems.
Studies like \cite{DBLP:journals/tbd/ChenLWLZ22} and \cite{DBLP:conf/kdd/WangFZWZA18} propose using discretized state transition graphs derived from trajectories to identify different driving behaviors.
In contrast, \cite{DBLP:conf/kdd/XuZZLZCX16} proposes to provide better driving behavior predictions by modeling the correlation between drivers’ skills and interactions hidden in their social networks.
Besides,  Next Point-of-Interest (POI) aims to recommend the next POIs drivers are most likely to visit based on their historical trajectories. 
Rao \etal \cite{rao2022graph} proposed a Spatial-Temporal Knowledge Graph (STKG), which can directly learn transition patterns between POIs. 
Trajectory prediction is similar to Next POI recommendation, which aims to predict the driver’s future visiting grid cell in the trajectory.
Xu \etal \cite{xu2022metaptp} designed a cluster-based~\cite{YangZWLXJ21} network initialization method based on a meta-learning algorithm to obtain initial personalized parameters for each trajectory. 
With the development of electric vehicle technologies, spatiotemporal data mining has also been applied to  electric vehicle tasks.
In order to help drivers find proper spots for charging, Zhang \etal \cite{zhang2021intelligent} proposed a framework called \textsc{Master} for charging station recommendation by considering each charging station as an individual agent and formulating the problem as a multi-objective multi-agent reinforcement learning task. To balance the use of charging stations, MAGC \cite{zhang2022multi} proposes to provide dynamic pricing for each charging request and achieve effective use of stations by formulating this problem as a mixed competitive-cooperative multi-agent reinforcement learning task with multiple long-term commercial goals.
\section{Conclusion}

\eat{
Vehicle energy consumption estimation is a challenging task related to various external and internal factors. However, the drivers are the most unstable variables, varying in individuals and road conditions. 
Additionally, we also introduced a meta-optimization technique to learn globally shared knowledge and enable the model to adapt quickly to every driver, 
alleviating cold-start problems and providing driver-specific VEC estimations. 
}

In this paper, we investigated the personalized vehicle energy consumption estimation problem by explicitly exploiting driving behaviors hidden in historical trajectories.
Specifically, we proposed a preference-aware meta-optimization framework (\baby) which consists of three major modules. 
We first proposed a driving preference learning module to extract latent
spatiotemporal preferences from historical trips.
After that, we constructed a selection-based driving behavior prediction module to estimate the possible driving behavior on a given route with the consideration of the driver's past relevant trips.
Furthermore, a driver-specific meta-optimization module is proposed to learn a shared global model parameter initialization that can be fast adapted to each long-tail driver with a few historical trips.
Extensive experiments on two large real-world datasets demonstrated the effectiveness of \baby against ten baselines.
In the future, we plan to deploy \baby to more cities so as to provide insightful information for various decision-making tasks such as individual trip planning and sustainable transportation system management.

\begin{acks}
    This research was supported in part by the National Natural Science Foundation of China under Grant No.62102110, Guangzhou Science and Technology Plan Guangzhou-HKUST(GZ) Joint Project No. 2023A03J0144, and Foshan HKUST Projects (FSUST21-FYTRI01A, FSUST21-FYTRI02A). 
\end{acks}

    

\bibliographystyle{ACM-Reference-Format}
\balance
\bibliography{refer}


\begin{thebibliography}{39}


\ifx \showCODEN    \undefined \def \showCODEN     #1{\unskip}     \fi
\ifx \showDOI      \undefined \def \showDOI       #1{#1}\fi
\ifx \showISBNx    \undefined \def \showISBNx     #1{\unskip}     \fi
\ifx \showISBNxiii \undefined \def \showISBNxiii  #1{\unskip}     \fi
\ifx \showISSN     \undefined \def \showISSN      #1{\unskip}     \fi
\ifx \showLCCN     \undefined \def \showLCCN      #1{\unskip}     \fi
\ifx \shownote     \undefined \def \shownote      #1{#1}          \fi
\ifx \showarticletitle \undefined \def \showarticletitle #1{#1}   \fi
\ifx \showURL      \undefined \def \showURL       {\relax}        \fi
\providecommand\bibfield[2]{#2}
\providecommand\bibinfo[2]{#2}
\providecommand\natexlab[1]{#1}
\providecommand\showeprint[2][]{arXiv:#2}

\bibitem[\protect\citeauthoryear{Al-Wreikat, Serrano, and Sodré}{Al-Wreikat
  et~al\mbox{.}}{2021}]%
        {al2021driving}
\bibfield{author}{\bibinfo{person}{Yazan Al-Wreikat}, \bibinfo{person}{Clara
  Serrano}, {and} \bibinfo{person}{José~Ricardo Sodré}.}
  \bibinfo{year}{2021}\natexlab{}.
\newblock \showarticletitle{Driving behaviour and trip condition effects on the
  energy consumption of an electric vehicle under real-world driving}.
\newblock \bibinfo{journal}{\emph{Applied Energy}}  \bibinfo{volume}{297}
  (\bibinfo{year}{2021}), \bibinfo{pages}{117096}.
\newblock
\showISSN{0306-2619}


\bibitem[\protect\citeauthoryear{Chen, Liu, Wang, Liao, and Zhang}{Chen
  et~al\mbox{.}}{2022b}]%
        {DBLP:journals/tbd/ChenLWLZ22}
\bibfield{author}{\bibinfo{person}{Chao Chen}, \bibinfo{person}{Qiang Liu},
  \bibinfo{person}{Xingchen Wang}, \bibinfo{person}{Chengwu Liao}, {and}
  \bibinfo{person}{Daqing Zhang}.} \bibinfo{year}{2022}\natexlab{b}.
\newblock \showarticletitle{semi-Traj2Graph identifying fine-Grained driving
  style with {GPS} trajectory data via multi-task learning}.
\newblock \bibinfo{journal}{\emph{IEEE Transactions on Big Data}}
  \bibinfo{volume}{8}, \bibinfo{number}{6} (\bibinfo{year}{2022}),
  \bibinfo{pages}{1550--1565}.
\newblock


\bibitem[\protect\citeauthoryear{Chen, Lei, and Ukkusuri}{Chen
  et~al\mbox{.}}{2022a}]%
        {chen2022prediction}
\bibfield{author}{\bibinfo{person}{Xiaowei Chen}, \bibinfo{person}{Zengxiang
  Lei}, {and} \bibinfo{person}{Satish~V. Ukkusuri}.}
  \bibinfo{year}{2022}\natexlab{a}.
\newblock \showarticletitle{Prediction of road-level energy consumption of
  battery electric vehicles}. In \bibinfo{booktitle}{\emph{Proceedings of the
  25th {IEEE} {ITSC} International Conference on Intelligent Transportation
  Systems, October 8-12, 2022}}. \bibinfo{publisher}{{IEEE}},
  \bibinfo{pages}{2550--2555}.
\newblock


\bibitem[\protect\citeauthoryear{Chen, Zhang, and Sun}{Chen
  et~al\mbox{.}}{2021}]%
        {chen2021data}
\bibfield{author}{\bibinfo{person}{Yuche Chen}, \bibinfo{person}{Yunteng
  Zhang}, {and} \bibinfo{person}{Ruixiao Sun}.}
  \bibinfo{year}{2021}\natexlab{}.
\newblock \showarticletitle{Data-driven estimation of energy consumption for
  electric bus under real-world driving conditions}.
\newblock \bibinfo{journal}{\emph{Transportation Research Part D: Transport and
  Environment}}  \bibinfo{volume}{98} (\bibinfo{year}{2021}),
  \bibinfo{pages}{102969}.
\newblock
\showISSN{1361-9209}


\bibitem[\protect\citeauthoryear{Chung, G{\"{u}}l{\c{c}}ehre, Cho, and
  Bengio}{Chung et~al\mbox{.}}{2014}]%
        {chung2014empirical}
\bibfield{author}{\bibinfo{person}{Junyoung Chung},
  \bibinfo{person}{{\c{C}}aglar G{\"{u}}l{\c{c}}ehre},
  \bibinfo{person}{KyungHyun Cho}, {and} \bibinfo{person}{Yoshua Bengio}.}
  \bibinfo{year}{2014}\natexlab{}.
\newblock \showarticletitle{Empirical evaluation of gated recurrent neural
  networks on sequence modeling}.
\newblock \bibinfo{journal}{\emph{CoRR}}  \bibinfo{volume}{abs/1412.3555}
  (\bibinfo{year}{2014}).
\newblock
\showeprint[arXiv]{1412.3555}


\bibitem[\protect\citeauthoryear{De~Cauwer, Van~Mierlo, and
  Coosemans}{De~Cauwer et~al\mbox{.}}{2015}]%
        {de2015energy}
\bibfield{author}{\bibinfo{person}{Cedric De~Cauwer}, \bibinfo{person}{Joeri
  Van~Mierlo}, {and} \bibinfo{person}{Thierry Coosemans}.}
  \bibinfo{year}{2015}\natexlab{}.
\newblock \showarticletitle{Energy consumption prediction for electric vehicles
  based on real-world data}.
\newblock \bibinfo{journal}{\emph{Energies}} \bibinfo{volume}{8},
  \bibinfo{number}{8} (\bibinfo{year}{2015}), \bibinfo{pages}{8573--8593}.
\newblock
\showISSN{1996-1073}


\bibitem[\protect\citeauthoryear{De~Cauwer, Verbeke, Coosemans, Faid, and
  Van~Mierlo}{De~Cauwer et~al\mbox{.}}{2017}]%
        {de2017data}
\bibfield{author}{\bibinfo{person}{Cedric De~Cauwer}, \bibinfo{person}{Wouter
  Verbeke}, \bibinfo{person}{Thierry Coosemans}, \bibinfo{person}{Saphir Faid},
  {and} \bibinfo{person}{Joeri Van~Mierlo}.} \bibinfo{year}{2017}\natexlab{}.
\newblock \showarticletitle{A Data-driven method for energy consumption
  prediction and energy-efficient routing of electric vehicles in real-world
  conditions}.
\newblock \bibinfo{journal}{\emph{Energies}} \bibinfo{volume}{10},
  \bibinfo{number}{5} (\bibinfo{year}{2017}).
\newblock
\showISSN{1996-1073}


\bibitem[\protect\citeauthoryear{Ding, Chen, Zhang, Guo, Yu, and Wang}{Ding
  et~al\mbox{.}}{2017}]%
        {ding2017greenplanner}
\bibfield{author}{\bibinfo{person}{Yan Ding}, \bibinfo{person}{Chao Chen},
  \bibinfo{person}{Shu Zhang}, \bibinfo{person}{Bin Guo},
  \bibinfo{person}{Zhiwen Yu}, {and} \bibinfo{person}{Yasha Wang}.}
  \bibinfo{year}{2017}\natexlab{}.
\newblock \showarticletitle{GreenPlanner: Planning personalized fuel-efficient
  driving routes using multi-sourced urban data}. In
  \bibinfo{booktitle}{\emph{Proceedings of the 2017 {IEEE} {Percom}
  International Conference on Pervasive Computing and Communications, March
  13-17, 2017}}. \bibinfo{publisher}{{IEEE} Computer Society},
  \bibinfo{pages}{207--216}.
\newblock


\bibitem[\protect\citeauthoryear{Dosovitskiy, Beyer, Kolesnikov, Weissenborn,
  Zhai, Unterthiner, Dehghani, Minderer, Heigold, Gelly, Uszkoreit, and
  Houlsby}{Dosovitskiy et~al\mbox{.}}{2021}]%
        {dosovitskiy2020image}
\bibfield{author}{\bibinfo{person}{Alexey Dosovitskiy}, \bibinfo{person}{Lucas
  Beyer}, \bibinfo{person}{Alexander Kolesnikov}, \bibinfo{person}{Dirk
  Weissenborn}, \bibinfo{person}{Xiaohua Zhai}, \bibinfo{person}{Thomas
  Unterthiner}, \bibinfo{person}{Mostafa Dehghani}, \bibinfo{person}{Matthias
  Minderer}, \bibinfo{person}{Georg Heigold}, \bibinfo{person}{Sylvain Gelly},
  \bibinfo{person}{Jakob Uszkoreit}, {and} \bibinfo{person}{Neil Houlsby}.}
  \bibinfo{year}{2021}\natexlab{}.
\newblock \showarticletitle{An image is worth 16x16 words: Transformers for
  image recognition at scale}. In \bibinfo{booktitle}{\emph{Proceedings of the
  9th {ICCV} International Conference on Learning Representations, May 3-7,
  2021}}. \bibinfo{publisher}{OpenReview.net}.
\newblock


\bibitem[\protect\citeauthoryear{Elmi and Tan}{Elmi and Tan}{2021}]%
        {elmi2021deepfec}
\bibfield{author}{\bibinfo{person}{Sayda Elmi} {and}
  \bibinfo{person}{Kian{-}Lee Tan}.} \bibinfo{year}{2021}\natexlab{}.
\newblock \showarticletitle{DeepFEC: Energy consumption prediction under
  real-world driving conditions for smart cities}. In
  \bibinfo{booktitle}{\emph{Proceedings of the 2021 {ACM} {WWW} International
  World Wide Web Conference, April 19-23, 2021}}. \bibinfo{publisher}{{ACM} /
  {IW3C2}}, \bibinfo{pages}{1880--1890}.
\newblock


\bibitem[\protect\citeauthoryear{Finn, Abbeel, and Levine}{Finn
  et~al\mbox{.}}{2017}]%
        {finn2017model}
\bibfield{author}{\bibinfo{person}{Chelsea Finn}, \bibinfo{person}{Pieter
  Abbeel}, {and} \bibinfo{person}{Sergey Levine}.}
  \bibinfo{year}{2017}\natexlab{}.
\newblock \showarticletitle{Model-agnostic meta-learning for fast adaptation of
  deep networks}. In \bibinfo{booktitle}{\emph{Proceedings of the 34th {ICML}
  International Conference on Machine Learning, 6-11 August 2017}}
  \emph{(\bibinfo{series}{Proceedings of Machine Learning Research},
  Vol.~\bibinfo{volume}{70})}. \bibinfo{publisher}{{PMLR}},
  \bibinfo{pages}{1126--1135}.
\newblock


\bibitem[\protect\citeauthoryear{Hong, Lin, Yang, Li, Fu, Wang, Qie, and
  Ye}{Hong et~al\mbox{.}}{2020}]%
        {hong2020heteta}
\bibfield{author}{\bibinfo{person}{Huiting Hong}, \bibinfo{person}{Yucheng
  Lin}, \bibinfo{person}{Xiaoqing Yang}, \bibinfo{person}{Zang Li},
  \bibinfo{person}{Kun Fu}, \bibinfo{person}{Zheng Wang},
  \bibinfo{person}{Xiaohu Qie}, {and} \bibinfo{person}{Jieping Ye}.}
  \bibinfo{year}{2020}\natexlab{}.
\newblock \showarticletitle{HetETA: Heterogeneous information network embedding
  for estimating time of arrival}. In \bibinfo{booktitle}{\emph{Proceedings of
  the 26th {ACM} {SIGKDD} International Conference on Knowledge Discovery and
  Data Mining, August 23-27, 2020}}. \bibinfo{publisher}{{ACM}},
  \bibinfo{pages}{2444--2454}.
\newblock


\bibitem[\protect\citeauthoryear{Horn, Moor, Bock, Rieck, and Borgwardt}{Horn
  et~al\mbox{.}}{2020}]%
        {horn2020set}
\bibfield{author}{\bibinfo{person}{Max Horn}, \bibinfo{person}{Michael Moor},
  \bibinfo{person}{Christian Bock}, \bibinfo{person}{Bastian Rieck}, {and}
  \bibinfo{person}{Karsten~M. Borgwardt}.} \bibinfo{year}{2020}\natexlab{}.
\newblock \showarticletitle{Set functions for time series}. In
  \bibinfo{booktitle}{\emph{Proceedings of the 37th {ICML} International
  Conference on Machine Learning, 13-18 July 2020}}
  \emph{(\bibinfo{series}{Proceedings of Machine Learning Research},
  Vol.~\bibinfo{volume}{119})}. \bibinfo{publisher}{{PMLR}},
  \bibinfo{pages}{4353--4363}.
\newblock


\bibitem[\protect\citeauthoryear{Hua, Sevegnani, Yi, Birnie, and McAslan}{Hua
  et~al\mbox{.}}{2022}]%
        {hua2022fine}
\bibfield{author}{\bibinfo{person}{Yining Hua}, \bibinfo{person}{Michele
  Sevegnani}, \bibinfo{person}{Dewei Yi}, \bibinfo{person}{Andrew Birnie},
  {and} \bibinfo{person}{Steve McAslan}.} \bibinfo{year}{2022}\natexlab{}.
\newblock \showarticletitle{Fine-grained {RNN} with transfer learning for
  energy consumption estimation on EVs}.
\newblock \bibinfo{journal}{\emph{IEEE Transactions on Industrial Informatics}}
  \bibinfo{volume}{18}, \bibinfo{number}{11} (\bibinfo{year}{2022}),
  \bibinfo{pages}{8182--8190}.
\newblock


\bibitem[\protect\citeauthoryear{Huang, Huang, Fang, Feng, Chen, Liu, Yuan, and
  Wang}{Huang et~al\mbox{.}}{2022}]%
        {huang2022dueta}
\bibfield{author}{\bibinfo{person}{Jizhou Huang}, \bibinfo{person}{Zhengjie
  Huang}, \bibinfo{person}{Xiaomin Fang}, \bibinfo{person}{Shikun Feng},
  \bibinfo{person}{Xuyi Chen}, \bibinfo{person}{Jiaxiang Liu},
  \bibinfo{person}{Haitao Yuan}, {and} \bibinfo{person}{Haifeng Wang}.}
  \bibinfo{year}{2022}\natexlab{}.
\newblock \showarticletitle{DuETA: Traffic congestion propagation pattern
  modeling via efficient graph learning for {ETA} prediction at Baidu Maps}. In
  \bibinfo{booktitle}{\emph{Proceedings of the 31st {ACM} {CIKM} International
  Conference on Information and Knowledge Management, October 17-21, 2022}}.
  \bibinfo{publisher}{{ACM}}, \bibinfo{pages}{3172--3181}.
\newblock


\bibitem[\protect\citeauthoryear{Liu, Liu, and Jiang}{Liu
  et~al\mbox{.}}{2022}]%
        {fliupractical2022}
\bibfield{author}{\bibinfo{person}{Fan Liu}, \bibinfo{person}{Hao Liu}, {and}
  \bibinfo{person}{Wenzhao Jiang}.} \bibinfo{year}{2022}\natexlab{}.
\newblock \showarticletitle{Practical Adversarial Attacks on Spatiotemporal
  Traffic Forecasting Models}. In \bibinfo{booktitle}{\emph{Proceedings of the
  2022 {NeurIPS} Annual Conference on Neural Information Processing Systems,
  Nov 28 - Dec 9, 2022}}.
\newblock


\bibitem[\protect\citeauthoryear{Liu, Peng, Yu, Wang, and Song}{Liu
  et~al\mbox{.}}{2021}]%
        {liu2021ldferr}
\bibfield{author}{\bibinfo{person}{Min Liu}, \bibinfo{person}{Zhaohui Peng},
  \bibinfo{person}{Xiaohui Yu}, \bibinfo{person}{Senzhang Wang}, {and}
  \bibinfo{person}{Qiao Song}.} \bibinfo{year}{2021}\natexlab{}.
\newblock \showarticletitle{LDFeRR: {A} fuel-efficient route recommendation
  approach for long-distance driving based on historical trajectories}. In
  \bibinfo{booktitle}{\emph{Proceedings of the 2021 {SIAM} {SDM} International
  Conference on Data Mining, April 29 - May 1, 2021}}.
  \bibinfo{publisher}{{SIAM}}, \bibinfo{pages}{73--81}.
\newblock


\bibitem[\protect\citeauthoryear{Moawad, Gurumurthy, Verbas, Li, Islam,
  Freyermuth, and Rousseau}{Moawad et~al\mbox{.}}{2021}]%
        {moawad2021deep}
\bibfield{author}{\bibinfo{person}{Ayman Moawad},
  \bibinfo{person}{Krishna~Murthy Gurumurthy}, \bibinfo{person}{Omer~{\"{O}}mer
  Verbas}, \bibinfo{person}{Zhijian Li}, \bibinfo{person}{Ehsan Islam},
  \bibinfo{person}{Vincent Freyermuth}, {and} \bibinfo{person}{Aymeric
  Rousseau}.} \bibinfo{year}{2021}\natexlab{}.
\newblock \showarticletitle{A deep learning approach for macroscopic energy
  consumption prediction with microscopic quality for electric vehicles}.
\newblock \bibinfo{journal}{\emph{CoRR}}  \bibinfo{volume}{abs/2111.12861}
  (\bibinfo{year}{2021}).
\newblock
\showeprint[arXiv]{2111.12861}


\bibitem[\protect\citeauthoryear{Oh, LeBlanc, and Peng}{Oh
  et~al\mbox{.}}{2022}]%
        {oh2020vehicle}
\bibfield{author}{\bibinfo{person}{Geunseob Oh}, \bibinfo{person}{David~J.
  LeBlanc}, {and} \bibinfo{person}{Huei Peng}.}
  \bibinfo{year}{2022}\natexlab{}.
\newblock \showarticletitle{Vehicle energy dataset (VED), {a} large-scale
  dataset for vehicle energy consumption research}.
\newblock \bibinfo{journal}{\emph{IEEE Transactions on Intelligent
  Transportation Systems}} \bibinfo{volume}{23}, \bibinfo{number}{4}
  (\bibinfo{year}{2022}), \bibinfo{pages}{3302--3312}.
\newblock


\bibitem[\protect\citeauthoryear{Ojeda, Chasse, and Goussault}{Ojeda
  et~al\mbox{.}}{2017}]%
        {ojeda2017fuel}
\bibfield{author}{\bibinfo{person}{Luis~Leon Ojeda}, \bibinfo{person}{Alexandre
  Chasse}, {and} \bibinfo{person}{Romain Goussault}.}
  \bibinfo{year}{2017}\natexlab{}.
\newblock \showarticletitle{Fuel consumption prediction for heavy-duty vehicles
  using digital maps}. In \bibinfo{booktitle}{\emph{Proceedings of the 20th
  {IEEE} {ICST} International Conference on Intelligent Transportation Systems,
  October 16-19, 2017}}. \bibinfo{publisher}{{IEEE}}, \bibinfo{pages}{1--7}.
\newblock


\bibitem[\protect\citeauthoryear{Perrotta, Parry, and Neves}{Perrotta
  et~al\mbox{.}}{2017}]%
        {perrotta2017application}
\bibfield{author}{\bibinfo{person}{Federico Perrotta}, \bibinfo{person}{Tony
  Parry}, {and} \bibinfo{person}{Lu{\'{\i}}s~C. Neves}.}
  \bibinfo{year}{2017}\natexlab{}.
\newblock \showarticletitle{Application of machine learning for fuel
  consumption modelling of trucks}. In \bibinfo{booktitle}{\emph{Proceedings of
  the 2017 {IEEE} {Big Data} International Conference on Big Data, December
  11-14, 2017}}. \bibinfo{publisher}{{IEEE} Computer Society},
  \bibinfo{pages}{3810--3815}.
\newblock


\bibitem[\protect\citeauthoryear{Petkevicius, Saltenis, Civilis, and
  Torp}{Petkevicius et~al\mbox{.}}{2021}]%
        {petkevicius2021probabilistic}
\bibfield{author}{\bibinfo{person}{Linas Petkevicius}, \bibinfo{person}{Simonas
  Saltenis}, \bibinfo{person}{Alminas Civilis}, {and} \bibinfo{person}{Kristian
  Torp}.} \bibinfo{year}{2021}\natexlab{}.
\newblock \showarticletitle{Probabilistic deep learning for electric-vehicle
  energy-use prediction}. In \bibinfo{booktitle}{\emph{Proceedings of the 17th
  {ACM} {SSTD} International Symposium on Spatial and Temporal Databases,
  August 23-25, 2021}}. \bibinfo{publisher}{{ACM}}, \bibinfo{pages}{85--95}.
\newblock


\bibitem[\protect\citeauthoryear{Rao, Chen, Liu, Shang, Yao, and Han}{Rao
  et~al\mbox{.}}{2022}]%
        {rao2022graph}
\bibfield{author}{\bibinfo{person}{Xuan Rao}, \bibinfo{person}{Lisi Chen},
  \bibinfo{person}{Yong Liu}, \bibinfo{person}{Shuo Shang},
  \bibinfo{person}{Bin Yao}, {and} \bibinfo{person}{Peng Han}.}
  \bibinfo{year}{2022}\natexlab{}.
\newblock \showarticletitle{Graph-flashback network for next location
  recommendation}. In \bibinfo{booktitle}{\emph{Proceedings of the 28th {ACM}
  {SIGKDD} International Conference on Knowledge Discovery and Data Mining,
  August 14 - 18, 2022}}. \bibinfo{publisher}{{ACM}},
  \bibinfo{pages}{1463--1471}.
\newblock


\bibitem[\protect\citeauthoryear{Rauh, Franke, and Krems}{Rauh
  et~al\mbox{.}}{2015}]%
        {rauh2015understanding}
\bibfield{author}{\bibinfo{person}{Nadine Rauh}, \bibinfo{person}{Thomas
  Franke}, {and} \bibinfo{person}{Josef~F. Krems}.}
  \bibinfo{year}{2015}\natexlab{}.
\newblock \showarticletitle{Understanding the impact of electric vehicle
  driving experience on range anxiety}.
\newblock \bibinfo{journal}{\emph{Human Factors: The Journal of Human Factors
  and Ergonomic Society}} \bibinfo{volume}{57}, \bibinfo{number}{1}
  (\bibinfo{year}{2015}), \bibinfo{pages}{177--187}.
\newblock


\bibitem[\protect\citeauthoryear{Roy, Nambi, Sobti, Ganu, Kalyanaraman, Akella,
  Devi, and Sundaresan}{Roy et~al\mbox{.}}{2022}]%
        {roy2022reliable}
\bibfield{author}{\bibinfo{person}{Millend Roy}, \bibinfo{person}{Akshay~Uttama
  Nambi}, \bibinfo{person}{Anupam Sobti}, \bibinfo{person}{Tanuja Ganu},
  \bibinfo{person}{Shivkumar Kalyanaraman}, \bibinfo{person}{Shankar Akella},
  \bibinfo{person}{Jaya~Subha Devi}, {and} \bibinfo{person}{S.~A. Sundaresan}.}
  \bibinfo{year}{2022}\natexlab{}.
\newblock \showarticletitle{Reliable energy consumption modeling for an
  electric vehicle fleet}. In \bibinfo{booktitle}{\emph{Proceedings of the 2022
  {ACM} {COMPASS} {SIGCAS/SIGCHI} Conference on Computing and Sustainable
  Societies, 29 June 2022 - 1 July 2022}}. \bibinfo{publisher}{{ACM}},
  \bibinfo{pages}{29--44}.
\newblock


\bibitem[\protect\citeauthoryear{Vaswani, Shazeer, Parmar, Uszkoreit, Jones,
  Gomez, Kaiser, and Polosukhin}{Vaswani et~al\mbox{.}}{2017}]%
        {vaswani2017attention}
\bibfield{author}{\bibinfo{person}{Ashish Vaswani}, \bibinfo{person}{Noam
  Shazeer}, \bibinfo{person}{Niki Parmar}, \bibinfo{person}{Jakob Uszkoreit},
  \bibinfo{person}{Llion Jones}, \bibinfo{person}{Aidan~N. Gomez},
  \bibinfo{person}{Lukasz Kaiser}, {and} \bibinfo{person}{Illia Polosukhin}.}
  \bibinfo{year}{2017}\natexlab{}.
\newblock \showarticletitle{Attention is all you need}. In
  \bibinfo{booktitle}{\emph{Proceedings of the 2017 {NeurIPS} Annual Conference
  on Neural Information Processing Systems, December 4-9, 2017}}.
  \bibinfo{pages}{5998--6008}.
\newblock


\bibitem[\protect\citeauthoryear{Wang, Zhao, Zhang, Luo, Qin, and Fang}{Wang
  et~al\mbox{.}}{2022b}]%
        {wang2022fine}
\bibfield{author}{\bibinfo{person}{Chenxing Wang}, \bibinfo{person}{Fang Zhao},
  \bibinfo{person}{Haichao Zhang}, \bibinfo{person}{Haiyong Luo},
  \bibinfo{person}{Yanjun Qin}, {and} \bibinfo{person}{Yuchen Fang}.}
  \bibinfo{year}{2022}\natexlab{b}.
\newblock \showarticletitle{Fine-grained trajectory-based travel time
  estimation for multi-city scenarios based on deep meta-learning}.
\newblock \bibinfo{journal}{\emph{IEEE Transactions on Intelligent
  Transportation Systems}} \bibinfo{volume}{23}, \bibinfo{number}{9}
  (\bibinfo{year}{2022}), \bibinfo{pages}{15716--15728}.
\newblock


\bibitem[\protect\citeauthoryear{Wang, Chen, Zhang, Wang, and Zhang}{Wang
  et~al\mbox{.}}{2019}]%
        {wang2019experience}
\bibfield{author}{\bibinfo{person}{Guang Wang}, \bibinfo{person}{Xiuyuan Chen},
  \bibinfo{person}{Fan Zhang}, \bibinfo{person}{Yang Wang}, {and}
  \bibinfo{person}{Desheng Zhang}.} \bibinfo{year}{2019}\natexlab{}.
\newblock \showarticletitle{Experience: Understanding long-term evolving
  patterns of shared electric vehicle networks}. In
  \bibinfo{booktitle}{\emph{Proceedings of the 25th {ACM} {MobiCom} Annual
  International Conference on Mobile Computing and Networking, MobiCom 2019,
  October 21-25, 2019}}. \bibinfo{publisher}{{ACM}},
  \bibinfo{pages}{9:1--9:12}.
\newblock


\bibitem[\protect\citeauthoryear{Wang, Fu, Zhang, Wang, Zheng, and
  Aggarwal}{Wang et~al\mbox{.}}{2018b}]%
        {DBLP:conf/kdd/WangFZWZA18}
\bibfield{author}{\bibinfo{person}{Pengyang Wang}, \bibinfo{person}{Yanjie Fu},
  \bibinfo{person}{Jiawei Zhang}, \bibinfo{person}{Pengfei Wang},
  \bibinfo{person}{Yu Zheng}, {and} \bibinfo{person}{Charu~C. Aggarwal}.}
  \bibinfo{year}{2018}\natexlab{b}.
\newblock \showarticletitle{You are how you drive: Peer and temporal-aware
  representation learning for driving behavior analysis}. In
  \bibinfo{booktitle}{\emph{Proceedings of the 24th {ACM} {SIGKDD}
  International Conference on Knowledge Discovery {\&} Data Mining, August
  19-23, 2018}}. \bibinfo{publisher}{{ACM}}, \bibinfo{pages}{2457--2466}.
\newblock


\bibitem[\protect\citeauthoryear{Wang, Fu, and Ye}{Wang et~al\mbox{.}}{2018a}]%
        {wang2018learning}
\bibfield{author}{\bibinfo{person}{Zheng Wang}, \bibinfo{person}{Kun Fu}, {and}
  \bibinfo{person}{Jieping Ye}.} \bibinfo{year}{2018}\natexlab{a}.
\newblock \showarticletitle{Learning to estimate the travel time}. In
  \bibinfo{booktitle}{\emph{Proceedings of the 24th {ACM} {SIGKDD}
  International Conference on Knowledge Discovery and Data Mining, August
  19-23, 2018}}. \bibinfo{publisher}{{ACM}}, \bibinfo{pages}{858--866}.
\newblock


\bibitem[\protect\citeauthoryear{Wang, Peng, Wang, and Song}{Wang
  et~al\mbox{.}}{2022a}]%
        {wang2022personalized}
\bibfield{author}{\bibinfo{person}{Zhan Wang}, \bibinfo{person}{Zhaohui Peng},
  \bibinfo{person}{Senzhang Wang}, {and} \bibinfo{person}{Qiao Song}.}
  \bibinfo{year}{2022}\natexlab{a}.
\newblock \showarticletitle{Personalized long-distance fuel-efficient route
  recommendation through historical trajectories mining}. In
  \bibinfo{booktitle}{\emph{Proceedings of the 15th {ACM} {WSDM} International
  Conference on Web Search and Data Mining, February 21 - 25, 2022}}.
  \bibinfo{publisher}{{ACM}}, \bibinfo{pages}{1072--1080}.
\newblock


\bibitem[\protect\citeauthoryear{Xing, Lv, Cao, and Lu}{Xing
  et~al\mbox{.}}{2020}]%
        {xing2020energy}
\bibfield{author}{\bibinfo{person}{Yang Xing}, \bibinfo{person}{Chen Lv},
  \bibinfo{person}{Dongpu Cao}, {and} \bibinfo{person}{Chao Lu}.}
  \bibinfo{year}{2020}\natexlab{}.
\newblock \showarticletitle{Energy oriented driving behavior analysis and
  personalized prediction of vehicle states with joint time series modeling}.
\newblock \bibinfo{journal}{\emph{Applied Energy}}  \bibinfo{volume}{261}
  (\bibinfo{year}{2020}), \bibinfo{pages}{114471}.
\newblock
\showISSN{0306-2619}


\bibitem[\protect\citeauthoryear{Xu, Zhu, Zhao, Liu, Zhong, Chen, and Xiong}{Xu
  et~al\mbox{.}}{2016}]%
        {DBLP:conf/kdd/XuZZLZCX16}
\bibfield{author}{\bibinfo{person}{Tong Xu}, \bibinfo{person}{Hengshu Zhu},
  \bibinfo{person}{Xiangyu Zhao}, \bibinfo{person}{Qi Liu},
  \bibinfo{person}{Hao Zhong}, \bibinfo{person}{Enhong Chen}, {and}
  \bibinfo{person}{Hui Xiong}.} \bibinfo{year}{2016}\natexlab{}.
\newblock \showarticletitle{Taxi driving behavior analysis in latent
  vehicle-to-vehicle networks: {A} social influence perspective}. In
  \bibinfo{booktitle}{\emph{Proceedings of the 22nd {ACM} {SIGKDD}
  International Conference on Knowledge Discovery and Data Mining, August
  13-17, 2016}}. \bibinfo{publisher}{{ACM}}, \bibinfo{pages}{1285--1294}.
\newblock


\bibitem[\protect\citeauthoryear{Xu, Xu, Zhao, Zheng, Liu, Zhao, and Zhou}{Xu
  et~al\mbox{.}}{2022}]%
        {xu2022metaptp}
\bibfield{author}{\bibinfo{person}{Yuan Xu}, \bibinfo{person}{Jiajie Xu},
  \bibinfo{person}{Jing Zhao}, \bibinfo{person}{Kai Zheng}, \bibinfo{person}{An
  Liu}, \bibinfo{person}{Lei Zhao}, {and} \bibinfo{person}{Xiaofang Zhou}.}
  \bibinfo{year}{2022}\natexlab{}.
\newblock \showarticletitle{MetaPTP: An adaptive meta-optimized model for
  personalized spatial trajectory prediction}. In
  \bibinfo{booktitle}{\emph{Proceedings of the 28th {ACM} {SIGKDD}
  International Conference on Knowledge Discovery and Data Mining, August 14 -
  18, 2022}}. \bibinfo{publisher}{{ACM}}, \bibinfo{pages}{2151--2159}.
\newblock


\bibitem[\protect\citeauthoryear{Yang and Gid{\'{o}}falvi}{Yang and
  Gid{\'{o}}falvi}{2018}]%
        {Yang2018FastMM}
\bibfield{author}{\bibinfo{person}{Can Yang} {and}
  \bibinfo{person}{Gy{\"{o}}z{\"{o}} Gid{\'{o}}falvi}.}
  \bibinfo{year}{2018}\natexlab{}.
\newblock \showarticletitle{Fast map matching, an algorithm integrating hidden
  Markov model with precomputation}.
\newblock \bibinfo{journal}{\emph{International Journal of Geographical
  Information Science}} \bibinfo{volume}{32}, \bibinfo{number}{3}
  (\bibinfo{year}{2018}), \bibinfo{pages}{547--570}.
\newblock


\bibitem[\protect\citeauthoryear{Yang, Zhan, Wu, Liu, Xiong, and Jiang}{Yang
  et~al\mbox{.}}{2021}]%
        {YangZWLXJ21}
\bibfield{author}{\bibinfo{person}{Yang Yang}, \bibinfo{person}{De{-}Chuan
  Zhan}, \bibinfo{person}{Yi{-}Feng Wu}, \bibinfo{person}{Zhi{-}Bin Liu},
  \bibinfo{person}{Hui Xiong}, {and} \bibinfo{person}{Yuan Jiang}.}
  \bibinfo{year}{2021}\natexlab{}.
\newblock \showarticletitle{Semi-Supervised Multi-Modal Clustering and
  Classification with Incomplete Modalities}.
\newblock \bibinfo{journal}{\emph{{IEEE} Transactions on Knowledge and Data
  Engineering}} \bibinfo{volume}{33}, \bibinfo{number}{2}
  (\bibinfo{year}{2021}), \bibinfo{pages}{682--695}.
\newblock


\bibitem[\protect\citeauthoryear{Zhang, Liu, Han, Ge, and Xiong}{Zhang
  et~al\mbox{.}}{2022}]%
        {zhang2022multi}
\bibfield{author}{\bibinfo{person}{Weijia Zhang}, \bibinfo{person}{Hao Liu},
  \bibinfo{person}{Jindong Han}, \bibinfo{person}{Yong Ge}, {and}
  \bibinfo{person}{Hui Xiong}.} \bibinfo{year}{2022}\natexlab{}.
\newblock \showarticletitle{Multi-agent graph convolutional reinforcement
  learning for dynamic electric vehicle charging pricing}. In
  \bibinfo{booktitle}{\emph{Proceedings of the 28th {ACM} {SIGKDD}
  International Conference on Knowledge Discovery and Data Mining, August 14 -
  18, 2022}}. \bibinfo{publisher}{{ACM}}, \bibinfo{pages}{2471--2481}.
\newblock


\bibitem[\protect\citeauthoryear{Zhang, Liu, Wang, Xu, Xin, Dou, and
  Xiong}{Zhang et~al\mbox{.}}{2021}]%
        {zhang2021intelligent}
\bibfield{author}{\bibinfo{person}{Weijia Zhang}, \bibinfo{person}{Hao Liu},
  \bibinfo{person}{Fan Wang}, \bibinfo{person}{Tong Xu},
  \bibinfo{person}{Haoran Xin}, \bibinfo{person}{Dejing Dou}, {and}
  \bibinfo{person}{Hui Xiong}.} \bibinfo{year}{2021}\natexlab{}.
\newblock \showarticletitle{Intelligent electric vehicle charging
  recommendation based on multi-agent reinforcement learning}. In
  \bibinfo{booktitle}{\emph{Proceedings of the 2021 {ACM} {WWW} International
  World Wide Web Conference, April 19-23, 2021}}. \bibinfo{publisher}{{ACM} /
  {IW3C2}}, \bibinfo{pages}{1856--1867}.
\newblock


\bibitem[\protect\citeauthoryear{Zhou, Gou, Hu, Zhang, Xu, Jiang, Li, and
  Xiong}{Zhou et~al\mbox{.}}{2019}]%
        {zhou2019collaborative}
\bibfield{author}{\bibinfo{person}{Jingbo Zhou}, \bibinfo{person}{Shan Gou},
  \bibinfo{person}{Renjun Hu}, \bibinfo{person}{Dongxiang Zhang},
  \bibinfo{person}{Jin Xu}, \bibinfo{person}{Airong Jiang},
  \bibinfo{person}{Ying Li}, {and} \bibinfo{person}{Hui Xiong}.}
  \bibinfo{year}{2019}\natexlab{}.
\newblock \showarticletitle{A collaborative learning framework to tag
  refinement for points of interest}. In \bibinfo{booktitle}{\emph{Proceedings
  of the 25th {ACM} {SIGKDD} International Conference on Knowledge Discovery
  and Data Mining, {KDD} 2019, Anchorage, AK, USA, August 4-8, 2019}}.
  \bibinfo{pages}{1752--1761}.
\newblock


\end{thebibliography}
\clearpage
\appendix

\section{APPENDIX}

\subsection{Top-$K$ historical trip selection Algorithm}\label{assc:topk}

\begin{algorithm}[htb]
  \SetKwData{Left}{left}\SetKwData{This}{this}\SetKwData{Up}{up}
  \SetKwFunction{Union}{Union}\SetKwFunction{FindCompress}{FindCompress}
  \SetKwInOut{Input}{input}\SetKwInOut{Output}{output}

  \Input{The route of the target trip $R_c$ and its departure time $s_c$, and historical trips' routes $\{R_i\}^M_{i=1}$ and departure times $\{s_i\}^M_{i=1}$.}
  \Output{The target trip's most similar $K$ trips $H^K$.}
  \BlankLine
  \For{$i\leftarrow 1$ \KwTo $M$}{
    Calculate the route similarity by: $score^{route}_{c, i} = \enspace \mid R_c \cap R_i \mid$\;
    Calculate the temporal similarity by: $score^{time}_{c, i} = \enspace \mid s_c - s_i \mid$\;
  }
  \For{$i\leftarrow 1$ \KwTo $M$}{
    Normalize each similarity score as: $norm(score^*_{c, i}) = \frac{score^*_{c, i} - Min(score^*_c)}{Max(score^*_c) - Min(score^*_c)}$\;
    Calculate the final similarity score by: $score_{c, i} = norm(score^{route}_{c, i}) - norm(score^{time}_{c, i})$\;
  }
  Select $K$ historical trips with the highest scores as $H^K$.
  \caption{Top-$K$ historical trip selection}\label{alg:topk}
\end{algorithm}\DecMargin{1em}

\subsection{Meta-optimization Algorithm}\label{assc:meta}

\begin{algorithm}[htb]
  \SetKwData{Left}{left}\SetKwData{This}{this}\SetKwData{Up}{up}
  \SetKwFunction{Union}{Union}\SetKwFunction{FindCompress}{FindCompress}
  \SetKwInOut{Input}{input}\SetKwInOut{Output}{output}

  \Input{All drivers' task datasets $\{(D^u_s, D^u_q)\}^U_{u=1}$, inner loop learning rate $\eta$, outer loop learning rate $\gamma$, and fine-tuning learning rate $\omega$.}
  \Output{Model parameters $\{\theta^u\}^U_{u=1}$ adapted to every driver.}
  \BlankLine
  Random initialize model parameters $\theta$\;
  \For{$i\leftarrow 1$ \KwTo $N_{epoch}$}{
    \For{$u\leftarrow 1$ \KwTo $U$}{
      Calculate the gradient of loss on the support set as: $\nabla_\theta \mathcal{L}^i_{D^u_s}(f_\theta)$\;
      Update parameters with gradient descent as: $\theta^\prime \leftarrow \theta - \eta \nabla_\theta \mathcal{L}^i_{D_s^u}(f_\theta)$\;
      Calculate the loss on the query set as: $\mathcal{L}^i_{D^u_q}(f_{\theta^\prime})$\;
    }
    Update parameters with gradient descent as: $\theta \leftarrow \theta - \gamma \nabla_\theta \sum^U_{u=1} \mathcal{L}^i_{D_q^u}(f_{\theta^\prime})$\;
  }
  \For{$u\leftarrow 1$ \KwTo $U$}{
    Merge the support and query set as $D^u$\;
    Fine-tune the model by: $\theta^u \leftarrow \theta - \omega \nabla_\theta \mathcal{L}_{D^u}(f_\theta)$\;
  }
  \caption{Meta-optimization}\label{alg:meta}
\end{algorithm}\DecMargin{1em}

\eat{
\subsection{System Demo}
We have crafted a system demo to provide energy consumption estimation for given trips. Figure \ref{fig:deploytrip} and \ref{fig:deployroad} show screenshots of our demo system. It has two views. One is an overview of the trip, which presents the route, the origin and destination, and the estimated total energy consumption of the trip. Another is the road view, which can show the energy consumption of the road segment when the driver clicks it. Furthermore, this view can also present the road-level VEC with different colors (warm colors represent high energy consumption).

\begin{figure}[H]
\centering
\includegraphics[width=\columnwidth]{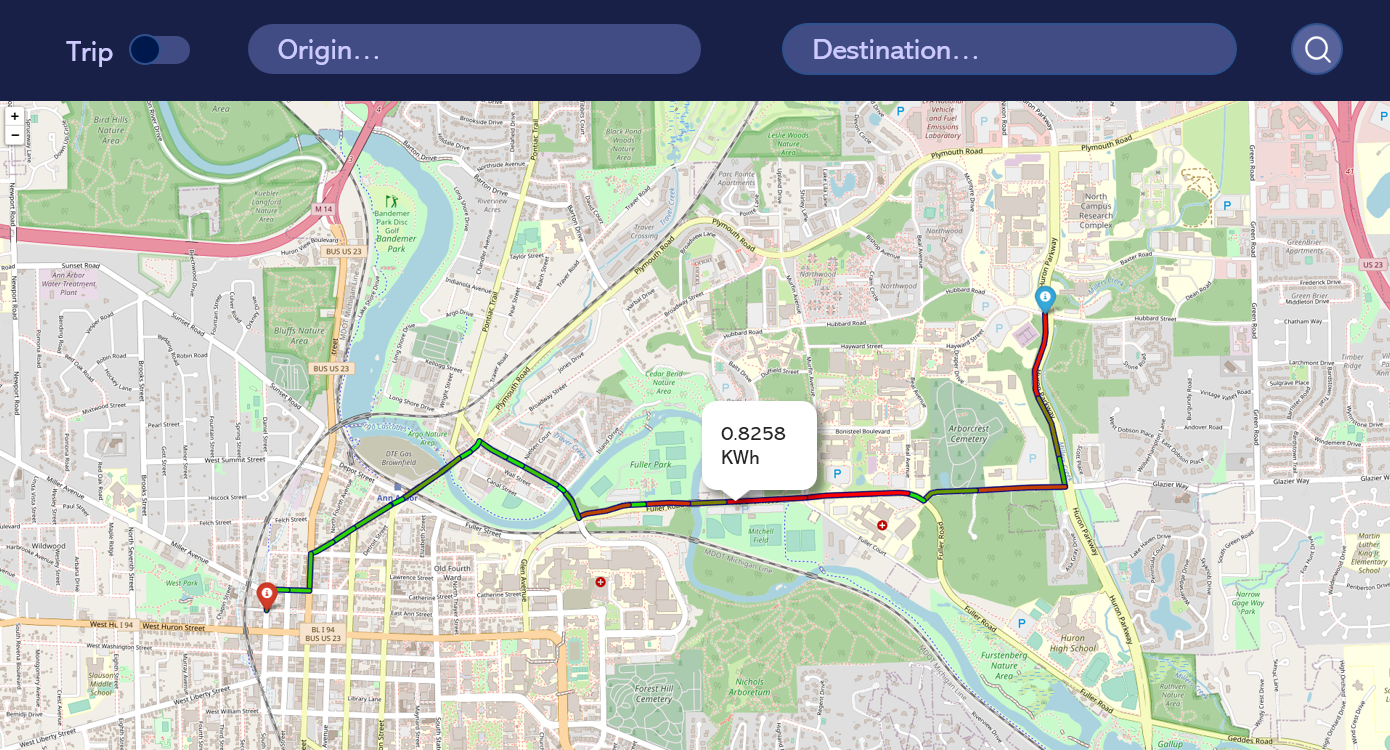}
\centering
\caption{System Demo: trip overview.}
\label{fig:deploytrip}
\end{figure}

\begin{figure}[H]
\centering
\includegraphics[width=\columnwidth]{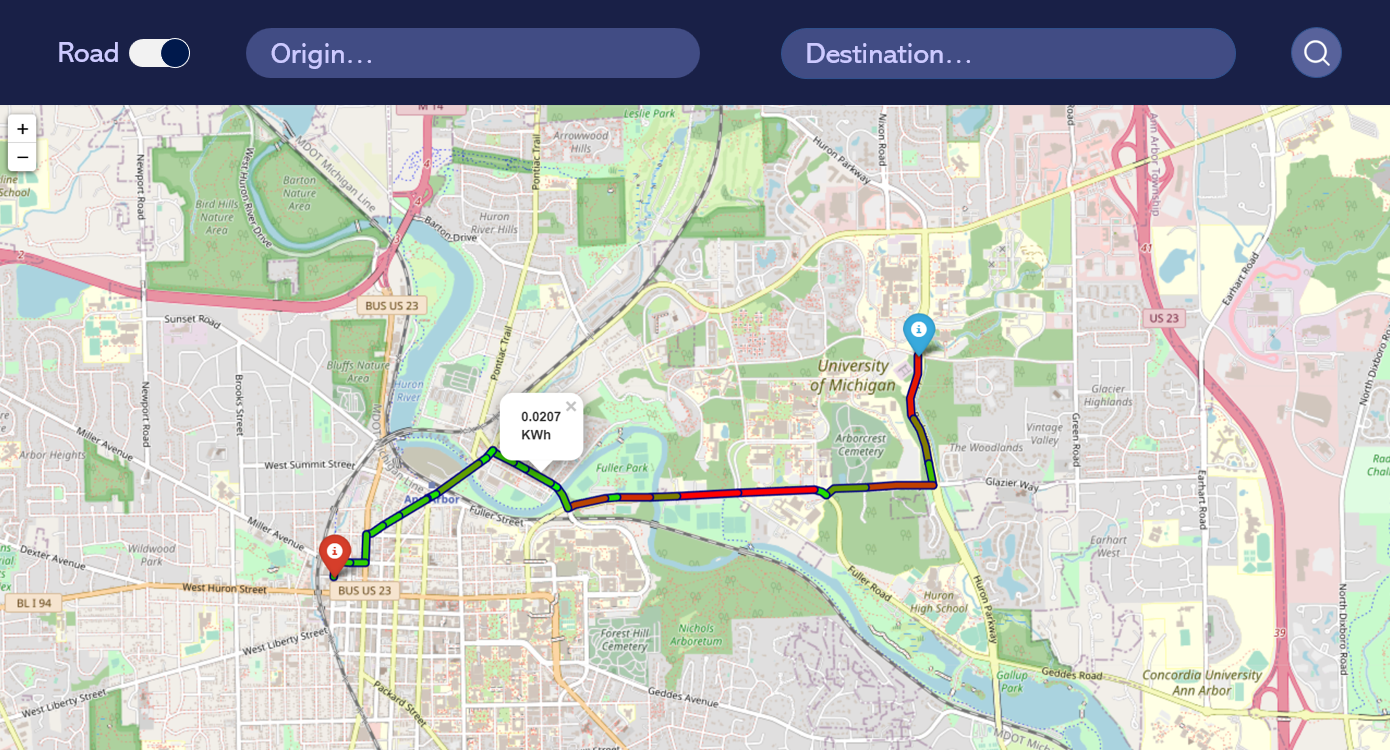}
\centering
\caption{System Demo: road view.}
\label{fig:deployroad}
\end{figure}
}

\subsection{Additional Case Study}
\begin{figure}[h]
\setlength{\abovecaptionskip}{0pt}
\setlength{\belowcaptionskip}{5pt}
  \begin{subfigure}{0.48\columnwidth}
    \centering
    \includegraphics[width=0.8\linewidth]{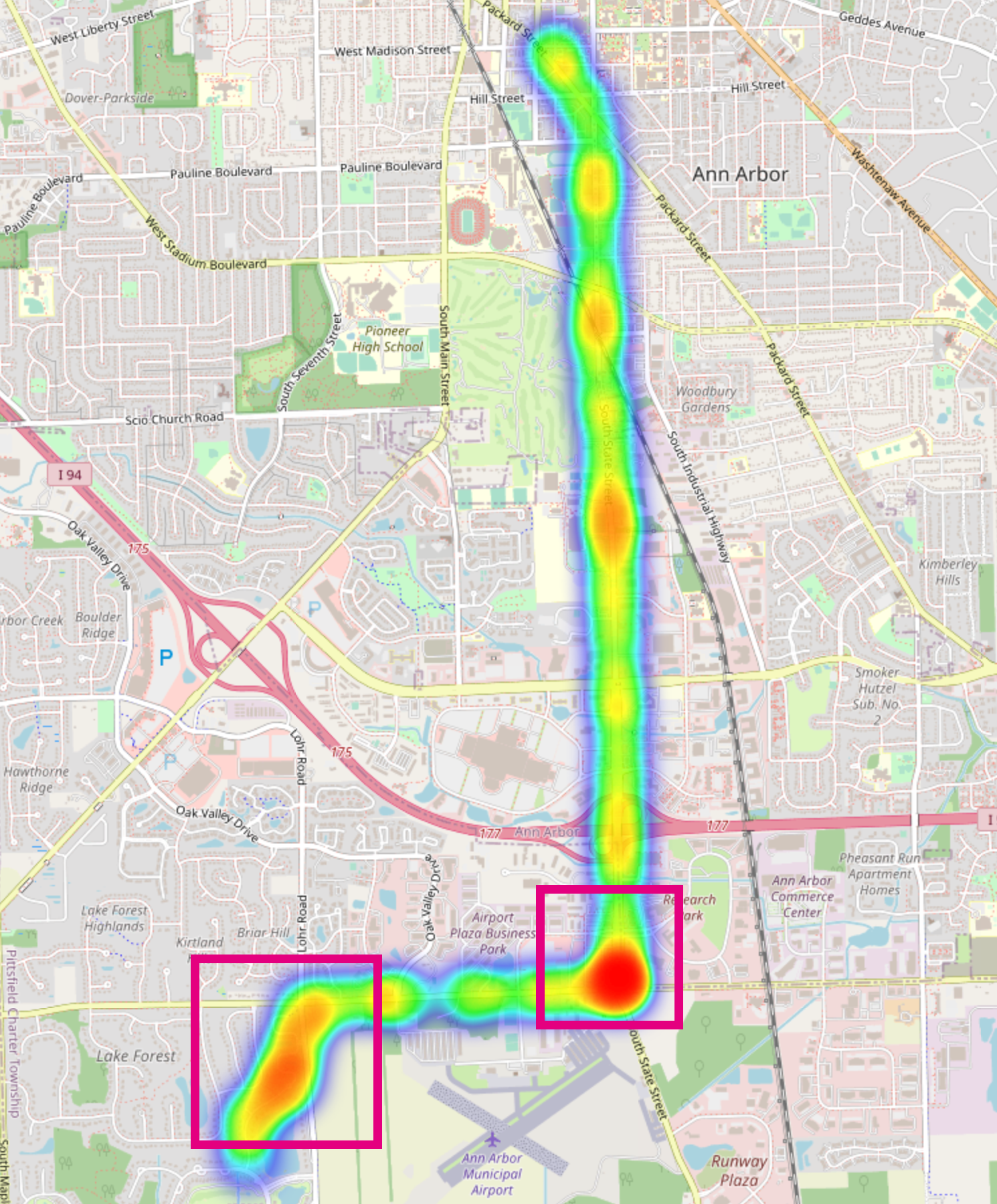}
    \caption{The ground truth VEC.}
    \label{fig:groundtruth_1}
  \end{subfigure}
  \begin{subfigure}{0.48\columnwidth}
    \centering
    \includegraphics[width=0.8\linewidth]{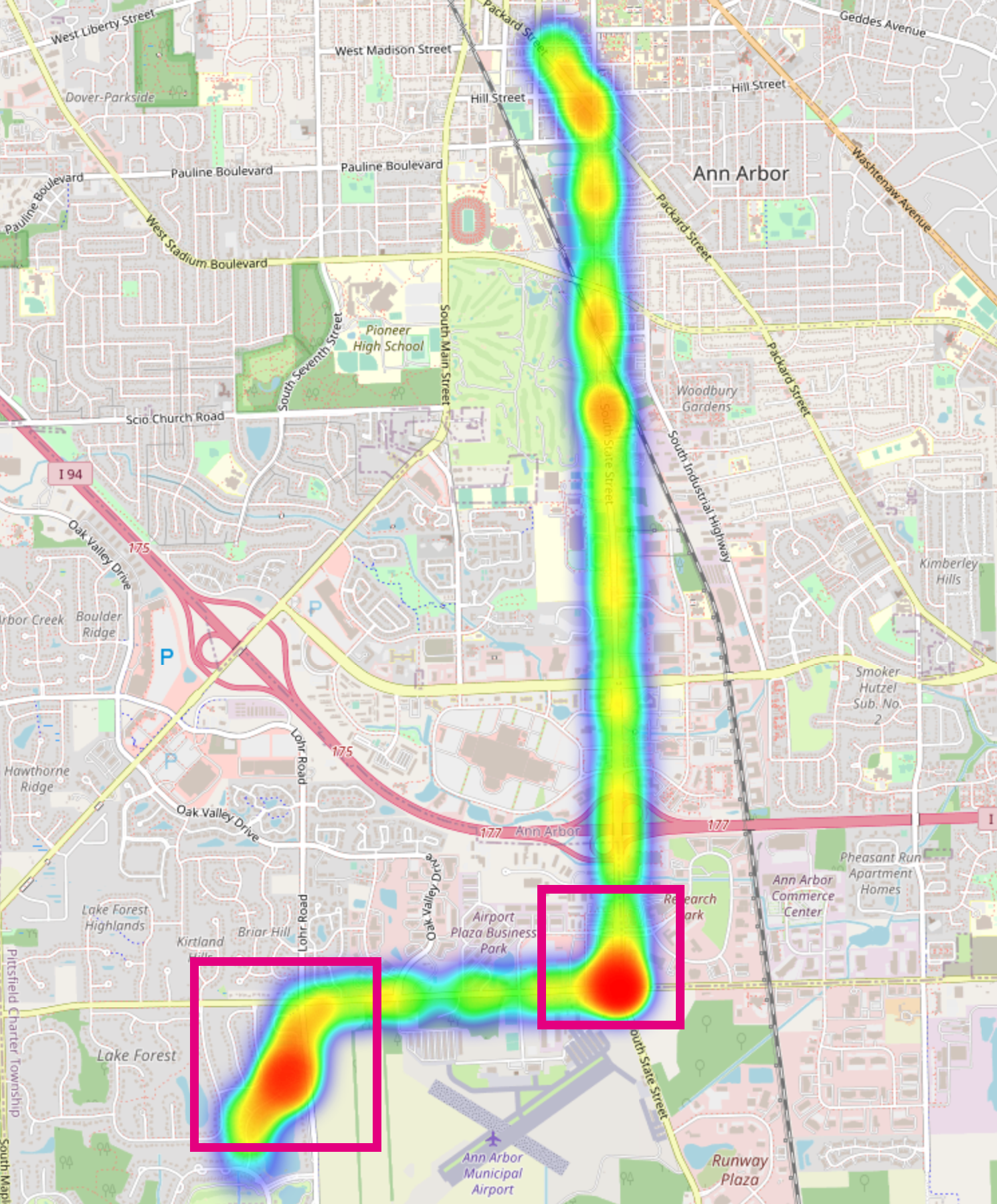}
    \caption{The predicted VEC.}
    \label{fig:pred_1}
  \end{subfigure}
  \caption{Additional case 1.}
  \label{fig:addcase_1}
\end{figure}

\begin{figure}[h]
\setlength{\abovecaptionskip}{0pt}
\setlength{\belowcaptionskip}{5pt}
  \begin{subfigure}{0.48\columnwidth}
    \centering
    \includegraphics[width=0.8\linewidth]{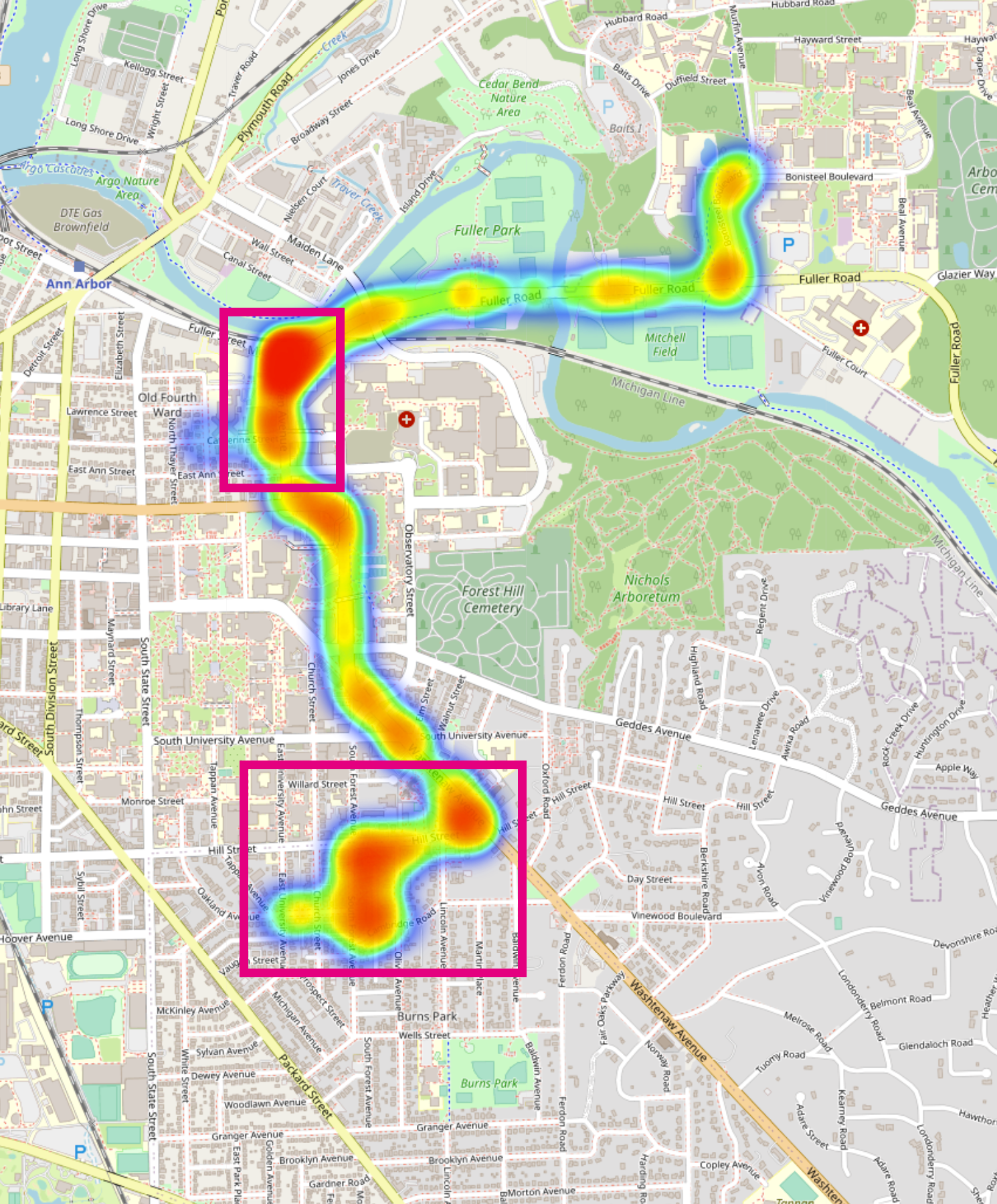}
    \caption{The ground truth VEC.}
    \label{fig:groundtruth_2}
  \end{subfigure}
  \begin{subfigure}{0.48\columnwidth}
    \centering
    \includegraphics[width=0.8\linewidth]{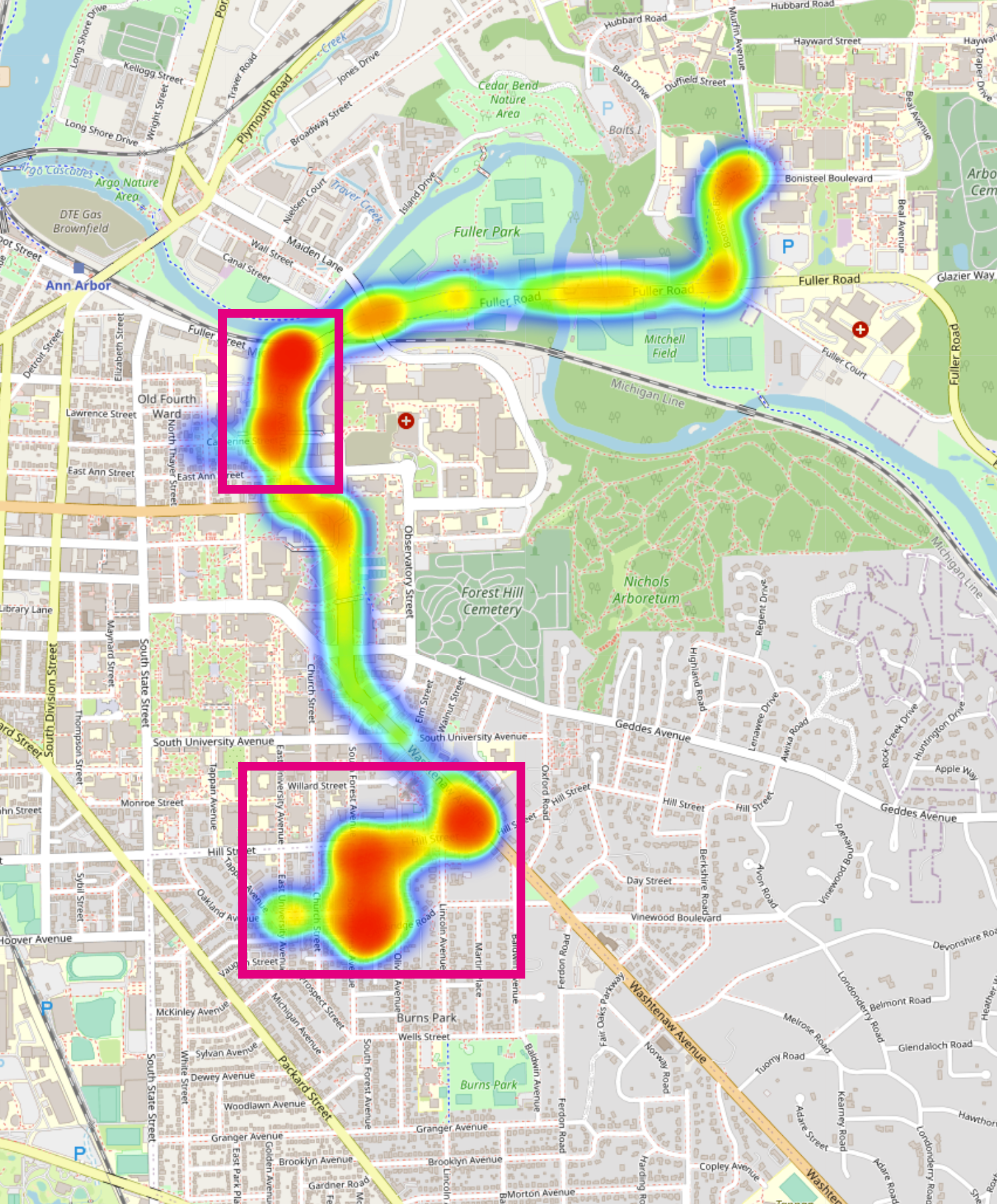}
    \caption{The predicted VEC.}
    \label{fig:pred_2}
  \end{subfigure}
  \caption{Additional case 2.}
  \label{fig:addcase_2}
\end{figure}

The ground truth and predicted VEC spatial distribution of additional cases are presented in Figure \ref{fig:addcase_1}-\ref{fig:addcase_2}.

\vspace{-0.5cm}

\vspace{-0.5cm}


\subsection{Prototype System}\label{subsec: prototype}
We have implemented a demo system to provide personalized energy consumption estimation for given trips. 
Figure~\ref{fig:deployroad} shows the screenshot of our demo system.
For each trip, the system displays the route, the origin and destination, and the estimated energy consumption of the trip, where warm colors indicate high energy consumption.
Moreover, the system also provides a road view to visualize the energy consumption of each road segment based on the past trajectories traversed.

\begin{figure}[H]
\centering
\includegraphics[width=\columnwidth]{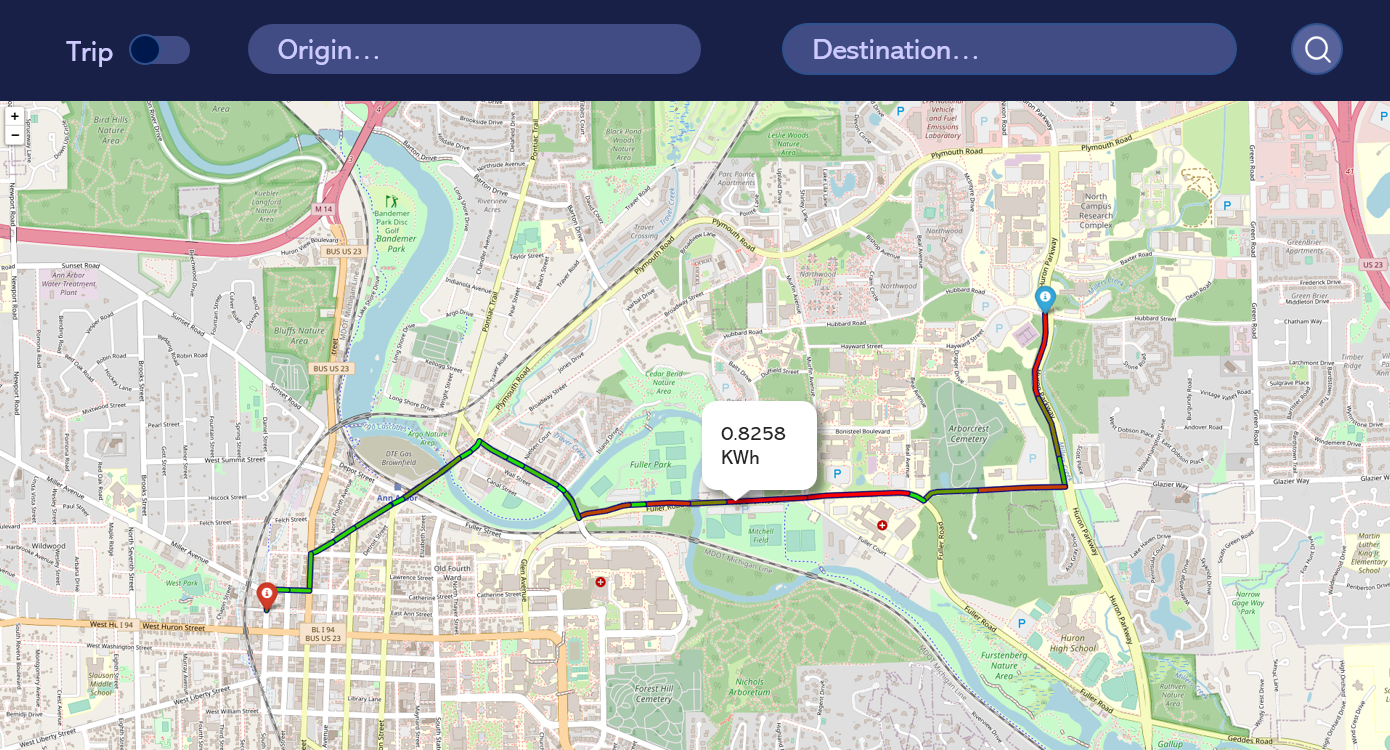}
\centering
\caption{Prototype system.}
\label{fig:deployroad}
\end{figure}


\end{document}